\documentclass[journal]{IEEEtran}

\usepackage{cite}
\usepackage{times}
\usepackage{soul}
\usepackage{url}
\usepackage[hidelinks]{hyperref}
\usepackage[utf8]{inputenc}
\usepackage[small]{caption}
\usepackage{graphicx}
\usepackage{amsmath}
\usepackage{amsthm}
\usepackage{amsfonts}
\usepackage{booktabs}
\usepackage{algorithm}
\usepackage{algorithmic}
\usepackage{subfigure}
\usepackage{color}
\usepackage{bm}
\usepackage{bbm}
\usepackage{comment}
\usepackage{multirow}

\newtheorem{theorem}{Theorem}
\newtheorem{corollary}{Corollary}
\usepackage{xspace}
\DeclareMathOperator*{\argmax}{arg\,max}
\DeclareMathOperator*{\argmin}{arg\,min}

\usepackage{array}
\usepackage{makecell}
\usepackage{enumitem}

\newcommand{\ADS}{\textsc{Ads}\xspace}

\begin{document}
\title{Taming Overconfident Prediction on Unlabeled Data from Hindsight}

\author{Jing~Li, 
        Yuangang~Pan,
        and~Ivor~W.~Tsang
\thanks{The authors are with the Australian Artificial Intelligence Institute (AAII), University of Technology Sydney, Australia. Email:  \href{mailto:Jing.Li-20@student.uts.edu.au}{Jing.Li-20@student.uts.edu.au}; {\{Yuangang.Pan, Ivor.Tsang\}@uts.edu.au}.}
}

%
%

\markboth{Journal of \LaTeX\ Class Files,~Vol.~14, No.~8, August~2015}%
{Shell \MakeLowercase{\textit{et al.}}: Bare Demo of IEEEtran.cls for IEEE Journals}
%



\maketitle

\begin{abstract}
Minimizing prediction uncertainty on unlabeled data is a key factor to achieve good performance in semi-supervised learning (SSL). The prediction uncertainty is typically expressed as the \emph{entropy} computed by the transformed probabilities in output space. Most existing works distill low-entropy prediction by either accepting the determining class (with the largest probability) as the true label or suppressing subtle predictions (with the smaller probabilities). 
Unarguably, these distillation strategies are usually heuristic and less informative for model training. From this discernment, this paper proposes a dual mechanism, named ADaptive Sharpening (\ADS), which first applies a soft-threshold to adaptively mask out determinate and negligible predictions, and then seamlessly sharpens the informed predictions, distilling certain predictions with the informed ones only. More importantly, we theoretically analyze the traits of \ADS by comparing with various distillation strategies. Numerous experiments verify that \ADS significantly improves the state-of-the-art SSL methods by making it a plug-in. Our proposed \ADS forges a cornerstone for future distillation-based SSL research.
\end{abstract}


%
\IEEEpeerreviewmaketitle

\section{Introduction}
\IEEEPARstart{L}{earning} with partially labeled data implicitly requires the model to exploit the missing labels on its own. This form of learning paradigm, known as semi-supervised learning~\cite{chapelle2006semi} (SSL), is of practical significance as labeling costs are now passed on to the subsequent algorithm design. 

The success of modern SSL algorithms is essentially attributed to two indispensable components designed for unlabeled data. (1) \emph{Consistency regularization}. It assumes that each unlabeled sample $\bm{u}$ should have the consistent prediction with its transformed counterpart $T(\bm{u})$. This self-supervised constraint on unlabeled data extends the conventional similarity regularization over neighboring samples~\cite{zhu2003semi,hadsell2006dimensionality,wang2020probabilistic}, providing an favourable optimization direction for classification, which thus is greatly boosted by recent data augmentation techniques~\cite{devries2017improved,cubuk2019autoaugment}. (2) \emph{Prediction distillation}. The predictions on unlabeled data are often of high uncertainty due to the absence of supervised information. Prediction distillation tries to enhance model prediction based on current output, e.g., following the principle of Minimum Entropy~\cite{grandvalet2005semi}. As a consequence, the decision boundary is encouraged to deviate from the region where the ambiguous predictions exist. Many SSL models~\cite{grandvalet2005semi,niu2014information,miyato2018virtual,oliver2018realistic} in literature have benefited from this component. 

Generally, given a labeled set $\mathcal{L}$ and an unlabeled set $\mathcal{U}$, a SSL model parameterized by $\bm{\theta}$ typically minimizes the following objective  
\begin{equation}\label{eq:J_total}
    \mathcal{J}(\mathcal{L},\mathcal{U}; \bm{\theta}) = \mathcal{J}_{\text{S}}+\alpha \mathcal{J}_{\text{C}}+\beta \mathcal{J}_{\text{D}},
\end{equation}
where $\mathcal{J}_{\text{S}}$ denotes a supervised loss on labeled set $\mathcal{L}$, $\mathcal{J}_{\text{C}}$ and $\mathcal{J}_{\text{D}}$ denote the consistency loss and distillation loss on unlabelled data $\mathcal{U}$, respectively. $\alpha$ and $\beta$ are non-negative weight factors for balancing three loss terms. Despite the great boost of consistency regularization to SSL, extra expertise is usually required to customize the effective data augmentation w.r.t. different data types~\cite{karras2020training}. Conversely, prediction distillation manages to learn from the model output, serving as a more general and practical tool in the machine learning community. Besides the aforementioned Minimum Entropy (ME), many other distillation strategies and their variants have been developed lately, such as Sharpening (SH)~\cite{xie2019unsupervised,berthelot2019mixmatch,berthelot2019remixmatch}, Pseudo-Labeling (PL)~\cite{lee2013pseudo,sohn2020fixmatch,arazo2020pseudo}, Negative Sampling (NS)~\cite{chen2020negative}. From the optimization perspective, although many of them progressively encourage partial unlabeled data to reach the low-entropy state, existing strategies unavoidably introduce the incorrect distillations because neural networks are argued not well calibrated~\cite{guo2017calibration}. The corresponding predictions are dubbed as \emph{overconfident predictions}\footnote{This problem is called over-confirmation for Pseudo-Labeling case in \cite{arazo2020pseudo}.} in this paper. That means for unlabeled predictions, not all distillations are beneficial.

Many recent studies can be interpreted as workarounds to this problem. By discriminating the importance of unlabeled data,~\cite{ren2020not} learns the sample-wise weights for unlabeled data through additive labelled validation set~\cite{ren2020not}. A more direct thought is leveraging the self-paced technique which selects the promising unlabeled samples~\cite{cascante2020curriculum} to do optimization in each iteration. Lately,~\cite{rizve2021defense} proposes to reduce the bias of PL by approximating the calibration criterion from the overall uncertainty. There methods practically require a lot of extra efforts, albeit some improvement are made. Specifically, the labelled validation set needed by~\cite{ren2020not} is hard to get in real world.  Self-paced strategy~\cite{cascante2020curriculum} may slow down the convergence rate, whose reported performance cannot rival the state-of-the-arts yet in their paper. The approximated calibration executed in~\cite{rizve2021defense} is implemented by Monte Carlo dropout~\cite{gal2016dropout} which demands multiple times inferences more than standard SSL training. Therefore, instead of following this branch of works, our research scope restricts to remedying overconfident predictions via tailor-designing more intelligent prediction distillation strategy which is expected to be less computational and more extendable.

\begin{figure*}[t] 
  \begin{minipage}[b]{0.68\textwidth}
  \scalebox{0.99}{
    \begin{tabular}{cccc}
    \toprule[1.3pt]
    \thead{Distillation Strategies} 
 & \thead{Candidates selection} & \thead{Activation \\in target label} & \thead{Uncertainty\\minimization} \\
\midrule
Minimum Entropy (ME) & None  & Single & \uppercase\expandafter{\romannumeral1}, \uppercase\expandafter{\romannumeral2}, \uppercase\expandafter{\romannumeral3} \\
Sharpening (SH) & None & Full   &  
\uppercase\expandafter{\romannumeral1}, \uppercase\expandafter{\romannumeral2}, \uppercase\expandafter{\romannumeral3} \\ 
Pseudo-Labeling (PL) & Hard threshold & Single  & \uppercase\expandafter{\romannumeral1} \\
Negative Sampling (NS) & Hard threshold  & Partial & \uppercase\expandafter{\romannumeral3}  \\ 
   \midrule
   ADaptive Sharpening (\ADS)  & Soft threshold   & Partial& \uppercase\expandafter{\romannumeral2} \\
   \bottomrule[1.3pt]
    \end{tabular}}
    \label{tb:mnist_limit_ep}
  \end{minipage}
  \hskip -0.2in
  \begin{minipage}[b]{0.28\textwidth}
  \includegraphics[width=\linewidth]{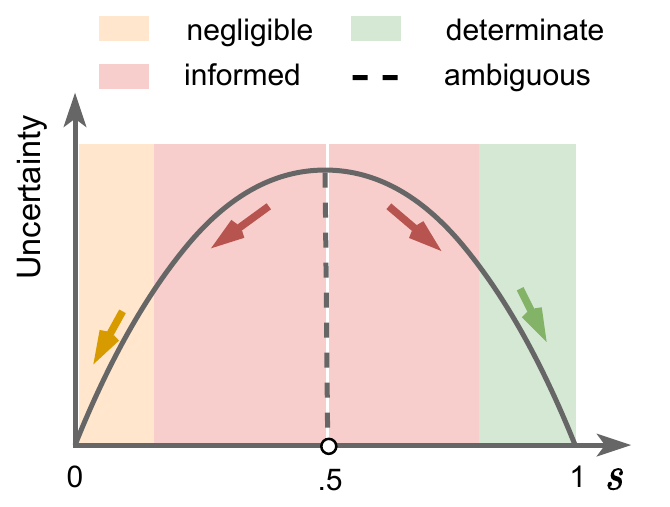}
  \vskip-0.7in
  \label{fig:uncertainty} 
  \end{minipage}
  \caption{\label{fig:distillation_intro} \textbf{Left:} Comparison among various prediction distillation strategies, each of which is viewed as a process of selecting candidate classes and aligning them with the categorical target label distribution. (\uppercase\expandafter{\romannumeral1}) enhance determinate predictions; (\uppercase\expandafter{\romannumeral2}) promote informed predictions; (\uppercase\expandafter{\romannumeral3}) suppress negligible predictions. \textbf{Right:} Different strategies turn out to have different ways to minimize prediction uncertainty. We showcase the binary classification for simplicity where $\bm{p} = (s, 1-s)$ where $0\le s \le 1$. }
\end{figure*}

Looking back on the execution of a SSL model which is equipped with a distillation component, we realize the overconfident prediction problem could be remedied if we stop greedily encouraging predictions to be one-hot and use the informative ones instead. With this \emph{hindsight}, we present ADaptive Sharpening (\ADS), a novel and simple prediction distillation method. By replacing the activation function with sparsemax transformation, \ADS performs adaptive selection on the logits layer, which prepares for the consequent loss computation. As a result, \ADS concentrates on a set of promising classes automatically selected for each unlabeled sample and avoids the unnecessary back-propagation for non-target class as well. Additionally, \ADS constructs a mild target label distribution by exponentially raising the sparse probabilities, leading to an informative optimization for minimizing prediction uncertainty. We borrow the name of Sharpening used in~\cite{Goodfellow-et-al-2016,berthelot2019mixmatch} to our method, but \ADS is essentially distinct from SH according to the comparison of Table~\ref{fig:distillation_intro}. We theoretically analyze the working mechanism of \ADS and demonstrate its superiority by numerous experiments. 
The major contributions of this paper are summarized as:
\begin{enumerate}[]
    \item We propose a new distillation strategy, named ADaptive  Sharpening  (\ADS), to distill low-entropy predictions for unlabeled data. The proposed \ADS is simply used a plug-in for SSL algorithms.
    \item  We analyze the connection and difference between \ADS and other distillation strategies through extensive comparisons. In particular, \ADS adaptively masks out the overconfident and negligible predictions and promotes the informed predictions only which is more instructive.
    \item We verify the superiority of our \ADS by combining it with advanced SSL models, including Virtual Adversarial Training (VAT) \cite{miyato2018virtual}, MixMatch \cite{berthelot2019mixmatch}, and FixMatch \cite{sohn2020fixmatch}. The experimental results on different benchmarks demonstrate \ADS achieves significant improvements over existing technologies. 
\end{enumerate}

This paper is organized as follows. In Section II, we present a series of recent SSL works that are closely related to our study. Section III revisits various distillation strategies which are analyzed from entropic view. The proposed \ADS and its corresponding SSL training model is introduced in Section IV, followed by the theoretical justification for its properties in Section V. Experiments in Section VI demonstrated the efficacy of \ADS. Section VII concludes the whole work and the last is the appendix.   

\section{Related Work}\label{previous work}
As there is a vast literature in SSL, we focus on the most related works which are expressed from the following two perspectives.

\textbf{Consistency Regularization} The concept of consistency can date back to the graph-based SSL work \cite{zhu2003semi} where similar training pairs are expected to have same label assignment. Recent SSL research has extended consistency to a more general case where natural unlabeled data and their transformed counterparts should correspond to the same output. This is closely related to the thought of self-supervised learning and many SSL studies~\cite{sajjadi2016regularization,xie2019unsupervised,berthelot2019remixmatch,han2020unsupervised} are furthering this direction. 
The adopted transformations play an important role therein, and they may include multiple passes of a single input with adversarial perturbations~\cite{miyato2015distributional}, randomized data augmentations~\cite{devries2017improved,cubuk2019autoaugment}, etc. Beyond the above consistency, using the prediction of weak augmentation to induce the prediction of corresponding strong augmentation~\cite{sohn2020fixmatch} lately shows superior performance. Summarily, consistency regularization is helpful because it facilitates the model to be insensitive to the possible disturbances, which is exactly desired for classification tasks.


\textbf{Prediction Distillation} 
Existing research approaching the certain predictions over unlabelled data typically adopts a post-processing distillation~\cite{DBLP:conf/nips/ChenKSNH20}. The pioneer strategy, such as ME~\cite{grandvalet2005semi} and its variant \cite{sajjadi2016regularization}, plainly punishes the high entropy predictions, enabling the model to have a decision boundary that traverses low-density regions. Thus, SSL models equipped with ME will exactly have one-hot prediction for all unlabelled data in theory. Differently, SH~\cite{Goodfellow-et-al-2016} constructs a mild target to align with in each step, showing a good performance in MixMatch \cite{berthelot2019mixmatch}. PL \cite{lee2013pseudo} is well-known and widely applied in recent research \cite{sohn2020fixmatch,cascante2020curriculum,arazo2020pseudo}. Opposite to above approaches, NS \cite{chen2020negative} lately resorts to minimizing the predictions of the classes the sample 
probably does not belong to. It is viewed as a safer option since searching for the negative classes based on the model output introduces less bias than directly picking up the true class.

\noindent\textbf{Notation} All the vectors are written in boldface. For a vector $\bm{p}$, the $k$-th entry is denoted as $p_i$, and $p_{(k)}$ is the $k$-th sorted coordinate of $\bm{p}$ in an descending order. The classification network with the model parameters of $\bm{\theta}$ is denoted by $f(;\bm{\theta})$, whose output logits vector is denoted by $\bm{z}$.

\section{Prediction Distillation Strategy Revisit}\label{sec:distillation_revisit}
In this section, we first present the formulation of existing distillation strategies and then analyze how they work from entropic view. 
\subsection{Formulation}
For a logits vector $\bm{z}$, softmax function produces strictly positive predictions by componentwise computing
\begin{equation}\label{eq:softmax}
    \text{softmax}_i(\bm{z}) =\frac{\text{exp}(z_i)}{\sum_j\text{exp}(z_j)}.
\end{equation}
For the better statement, we further denote $\bm{p} = \text{softmax}(\bm{z})$. Now we formulate every distillation strategy as follows.

\textbf{ME} The loss for minimum entropy of an unlabelled sample is written as
\begin{equation}
\mathcal{J}_{\text{D}}^{\text{ME}}(\bm{p}) = -\sum_{i=1}^{K}p_i\log p_i.
\end{equation}
Notice that ME is originally defined on unlabelled data, and it applies to augmented unlabelled data~\cite{miyato2018virtual} as well. 

\textbf{SH} Sharpening function~\cite{heaton2018ian,berthelot2019mixmatch} constructs the target label by adjusting the temperature of the current categorical distribution $\bm{p}$, 
\begin{equation}\label{eq:SH}
\text{SH}_i(\bm{p},\lambda) = \frac{p_i^{1/\lambda}}{\sum_j^{K}p_j^{1/\lambda}},
\end{equation}
where $\lambda$ ($\lambda>0$) is the temperature that controls how sharp the output distribution looks like. Then the corresponding distillation loss is formulated as 
\begin{equation}
    \mathcal{J}_{\text{D}}^{\text{SH}}(\bm{p})=\text{Dist}(\text{SH}(\bm{p},\lambda), \bm{p}).
\end{equation}

\textbf{PL} Naive Pseudo-Labeling~\cite{lee2013pseudo} could be described as a hard version of sharpening, and it picks the class which has the maximum predicted probability to be the target label, 
\begin{equation}
    \text{PL}_i(\bm{p})=\left\{
    \begin{array}{ll}
     1, & { \text{if} \, i=\argmax_i'p_{i}' }\\
     0, &  {\text{otherwise}}.
    \end{array}\right.
\end{equation}
In practice, a stricter condition applies~\cite{sohn2020fixmatch} that requires the maximum probability to exceed a predefined threshold $\tau_{\text{PL}}$. This means that PL does sample selection in each round optimization. For each selected sample, the distillation loss is written in a similar manner with SH, 
\begin{equation}
    \mathcal{J}_{\text{D}}^{\text{PL}}(\bm{p})=\text{Dist}(\text{PL}(\bm{p}), \bm{p}).
\end{equation}

\textbf{NS} Picking negative classes by a threshold $\tau_{\text{NS}}$ and minimizing their probabilities comes to the distillation loss of
\begin{equation}
\mathcal{J}_{\text{D}}^{\text{NS}}(\bm{p}) = -\log (1-\sum_{i=1}^{K} \mathbbm{I}(p_i<\tau_{\text{NS}})p_i),
\end{equation}
where $\mathbbm{I}(\cdot)$ is the indicator function. This loss term is proved to better approximate the true likelihood of unlabeled data in~\cite{chen2020negative} and we interpret it as one of the distillation baselines.

\subsection{Entropic View of Distillation}
We observe that existing distillation strategies are in pursuit of low-entropy predictions via post-processing the model predictions in different ways. According to their formulations, we summarize them into the left panel of Figure~\ref{fig:distillation_intro} where they are compared in terms of candidate classes selection, the sparsity of activation, and the way of uncertainty minimization. Before diving into more details, we need informally introduce the following definitions. 

The labeled data can be viewed as anchor points which initialize the model and induce the unlabeled prediction. Given \begin{small}$0 \! < \! \theta_1 \! \ll \! 0.5 \! \ll \! \theta_2 \! < \!1$\end{small}, we adhere the philosophy of this thought and partition the intact probability space into four intervals:
\begin{itemize}
\item \textbf{negligible:} \begin{small}$\{p_i|p_i<\theta_1, i= 1,2,\ldots, K\}$\end{small};
\item \textbf{ambiguous:} \begin{small}$\{p_i|p_i=\frac{1}{K}, i= 1,2,\ldots, K\}$\end{small};
\item \textbf{informed:} \begin{small}$\{p_i|\theta_1 \le p_i \le \theta_2,\ p_i\ne\frac{1}{K},\ i= 1,2,\ldots, K\}$\end{small};
\item \textbf{determinate:} \begin{small}$\{p_i|p_i>\theta_2, i= 1,2,\ldots, K\}$\end{small}.
\end{itemize}
The above definitions allow us to roughly categorize each prediction as one of the four types. Thus, we can see that ME and SH consider all classes as the candidates and aggressively minimize uncertainties over all categories except ambiguous ones which provide no useful information for distillation. By simply setting $\theta_1=\tau_{\text{NS}}$ and $\theta_2=\tau_{\text{PL}}$, we conclude that PL and NS resort to heuristic thresholds to ensure that distillation acts on relatively certain (determinate or negligible) predictions only. As a result, ME, SH, and PL suffer the problem of overconfident predictions~\cite{arazo2020pseudo,ren2020not} as they are plainly enhancing the determinate predictions (type \uppercase\expandafter{\romannumeral1} in Figure~\ref{fig:distillation_intro}). Notice that NS tries to penalize the negligible predictions, which seems to be safer. However, we demonstrate that it factually has the similar effect with PL for binary class classification as further discussed in Section~\ref{sec:inclined}.

For the sake of better understanding to the different distillation strategies, we display the four types of prediction in binary-class classification case on the right panel of Figure~\ref{fig:distillation_intro}. As we have interpreted all distillation strategies as uncertainty minimization, we indeed decompose each distillation strategy in this probability space. It is observed ambiguous prediction is special and it only takes a point. $\theta_1$ and $\theta_2$ does not coexist in existing methods, and they are not always empirically satisfying $\theta_1+\theta_2=1$. In this work, we aim to leverage the informed predictions only, which remedies the overconfident predictions by doing nothing to the extremely certain predictions (Extra benefits can be referred in Section~\ref{sec:EM}). Particularly, as in different training stages and for different unlabeled samples, the informed predictions should be relatively different and dynamic. Thus, the informed predictions are not quite aligned with the ones defined by hard thresholds used in PL and NS.


\section{\ADS Based SSL Framework}
\begin{figure}[t]
    \centering
    \includegraphics[width=0.45\textwidth]{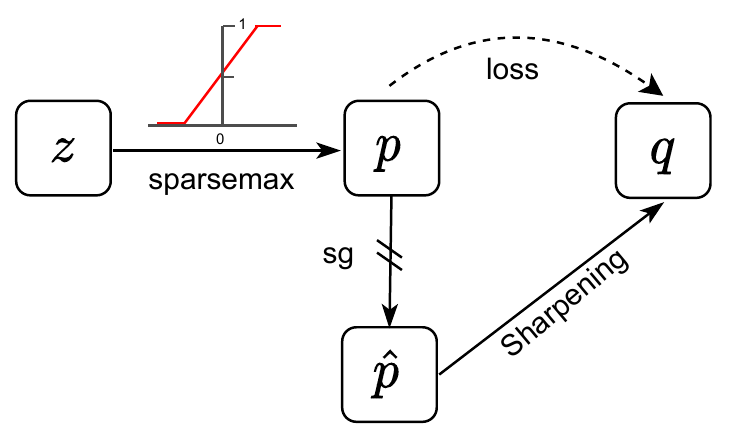}   
    \caption{\label{fig:pipeline}Distillation architecture of \ADS.}
\end{figure}

According to the understanding of prediction distillation indicated by the left panel of Figure~\ref{fig:distillation_intro}, we propose ADaptive Sharpening (\ADS), a dual distillation mechanism which only sharpens the informed predictions so as to avert the overconfident issues. An architecture of \ADS from logits to losses is showcased as Figure~\ref{fig:pipeline}. We also analyze how do other loss terms benefit from \ADS, which completes the presentation of concrete \ADS based SSL framework.   

\subsection{ADaptive Sharpening (\ADS)}
\subsubsection{Adaptive Selection via Sparsemax}\label{sec3.1}
Properly selecting partial classes to do distillation assists the classification model to concentrate on the most confusing classes for each unlabeled sample~\cite{chen2020negative}. Existing prediction distillation strategies, such as PL and NS, are plainly overlaid on the probabilistic distribution produced by softmax function, formulating a hard threshold-based post-processing approach \cite{DBLP:conf/nips/ChenKSNH20}. In spite of their popularity in recent research~\cite{sohn2020fixmatch,chen2020negative}, the major drawback of them is the adoption of a predefined threshold, which is not robust enough to account for individual variability and model variability at different training stages, dwindling its effectiveness in practical implementations.

Therefore, we aim to design a flexible sampler, which can automatically determine a partial set of promising classes as candidates for true class, being adaptive to each unlabelled sample at different training stages. To this end, we study this problem from neural networks' output, i.e., logits, and apply sparsemax \cite{martins2016softmax} to produce a sparse probability distribution directly, instead of applying a truncation over probability values afterward. 
For a neural networks' output logits vector $\bm{z}$, sparsemax is calculated by
\begin{equation}\label{eq:sparsemax}
    \text{sparsemax}_i(\bm{z})=[z_i-\tau(\bm{z})]_+,
\end{equation}
where $[u]_+:=\max(0,u)$. $\tau{(\bm{z})}$ serves as a soft threshold, which is briefly expressed as follows. Let $z_{(1)} \ge z_{(2)} \ge...\ge z_{(K)}$ be the sorted coordinates of $\bm{z}$, and define $k(\bm{z}):=\max\{k\in\{1,...,K\}|1+k z_{(k)}>\sum_{j\le k}z_{(j)}\}$. Then, $\tau(\bm{z}) = \frac{(\sum_{j\le k(\bm{z})}z_{(j)})-1}{k(\bm{z})}$. Basically, given a logits vector $\bm{z}$, sparsemax finds a threshold $\tau_{(\bm{z})}$ and sets probabilities below $\tau_{(\bm{z})}$ to zeros while satisfying $\sum_j[z_j-\tau(\bm{z})]_+=1$. $k(\bm{z})$ indicates the support of $\text{sparsemax}(\bm{z})$ which makes $\tau_{(\bm{z})}$ fall in the ``maximum gap" of the sorted logit values. We are in favor of this valuable trait in SSL as sparsemax adaptively rules out some unrelated classes, which is viewed as more flexible and less biased compared with those using a hard threshold.

Selecting proper candidates motivates the employment of sparsemax. This process also implies the informed predictions desired by our distillation component. We focus on the pipeline of our method here and leave the analyses how \ADS fulfill our hindsight in Section~\ref{sec:inclined}. 

\subsubsection{Sharpening on Sparse Probabilities}\label{sec3.2}

Prediction distillation acts as a self-training process. To iteratively refine the classification model, we define the distillation loss of \ADS as a distance, e.g., Kullback-Leibler (KL) divergence, between the raw prediction $\bm{p}$ and the auxiliary target label distribution $\bm{q}$:
\begin{equation}\label{eq:J_d}
    \mathcal{J}_{\text{D}} = \sum_{\bm{u} \in \mathcal{U}}\text{KL}(\bm{q}||\bm{p})=\sum_{\bm{u} \in \mathcal{U}}\sum_iq_i\log \frac{q_i}{p_i},
\end{equation}
where $p_i = \text{sparsemax}_i(f(\bm{u};\bm{\theta}))$. The choice of target label distribution $\bm{q}$ is crucial for \ADS's performance. Particularly, we would like our target label distribution $\bm{q}$ to have the following properties: (1) strengthen confident predictions, (2) exclude the contributions of negative class predictions.

According to the comparison in Figure~\ref{fig:distillation_intro}, Sharpening is a feasible option to satisfy the above requirements. Having a sparse prediction $\bm{p}$ in hand, to construct the according target label distribution $\bm{q}$, we first exert $\hat{\bm{p}}=\text{sg}(\bm{p})$, where $\text{sg}$ stands for the stop-gradient operator that is defined as identity during forward computation and has zero partial derivatives. After that, similar to Eq.~\eqref{eq:SH}, we calculate every coordinate $q_i$ of $\bm{q}$ as a normalized probability raised from ${\hat{p}}_i$ with a power of $r$:
\begin{equation}
    q_i = \frac{{\hat{p}}_i^r}{\sum_j {\hat{p}}_j^r}.
\end{equation}
Obviously, if ${\hat{p}}_i=0$, we have $q_i=0$, and if $\bm{\hat{p}}=\frac{1}{K}\bm{1}$, we have $\bm{q}=\bm{\hat{p}}$. 
Otherwise, as $r \rightarrow \infty $, the categorical distribution $\bm{q}$ will approach a delta distribution. We emphasize that although \ADS borrows the idea of SH for constructing the target label distribution, they have the different functionalities during distillation. Particularly, we will show \ADS does not aggressively distill relatively certain predictions in Section~\ref{sec:inclined}, which mitigates the overconfident risk.

Figure~\ref{fig:pipeline} shows the process of distillation design starting from derived logits. Note that although Eq.~\eqref{eq:J_d} is defined on unlabeled data, it could be extended to augmented unlabeled instances in some methods. Accordingly, the augmented data will pass the sparsemax once they are expected to participate the distillation loss.

\subsection{In Conjunction with Other Loss}
\begin{figure*}[t]
\centering     
\subfigure[Target Probability]{\label{fig:a}\includegraphics[width=52.5mm]{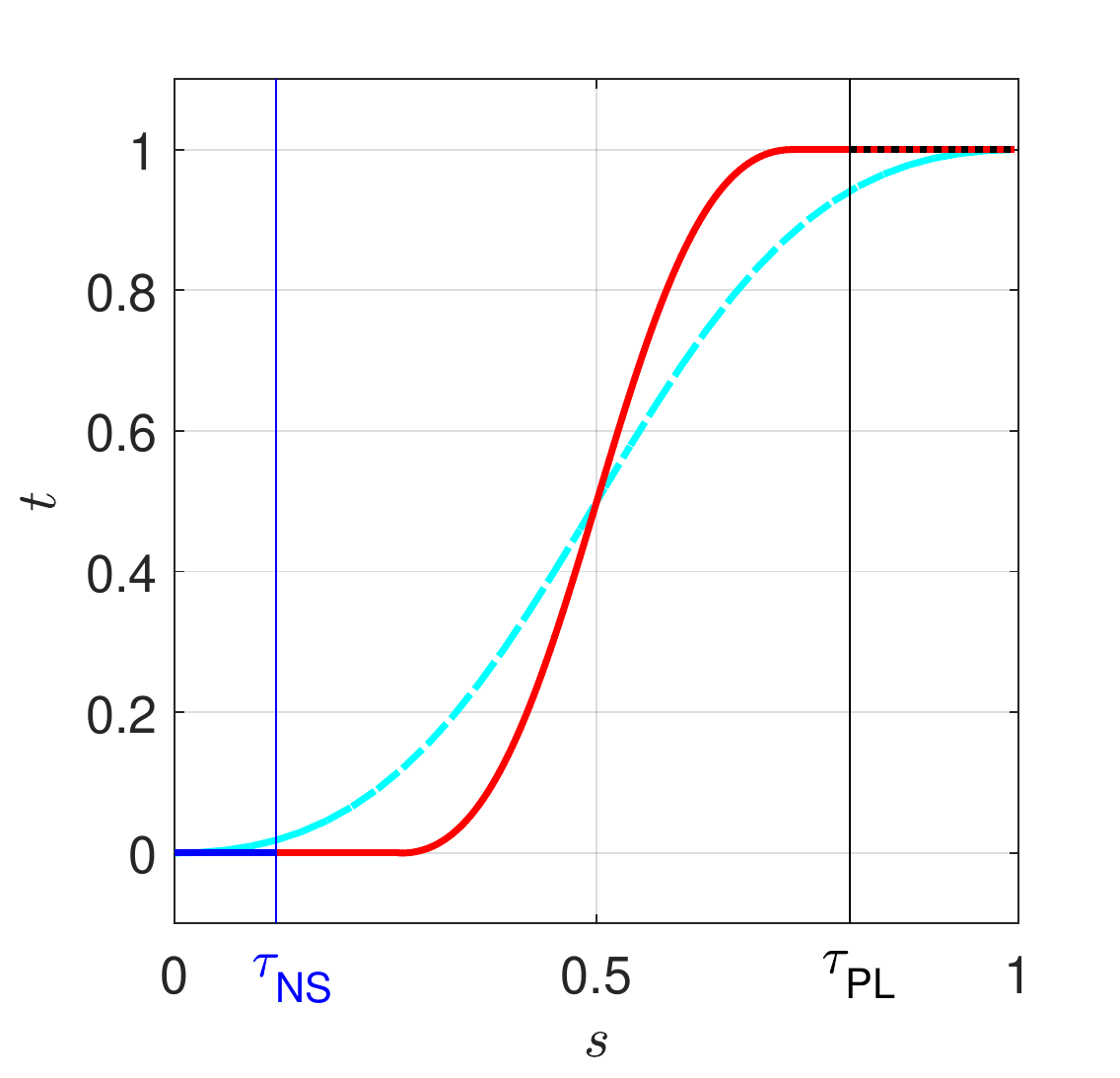}}
\hfill
\subfigure[Distillation Loss]{\label{fig:b}\includegraphics[width=51.5mm]{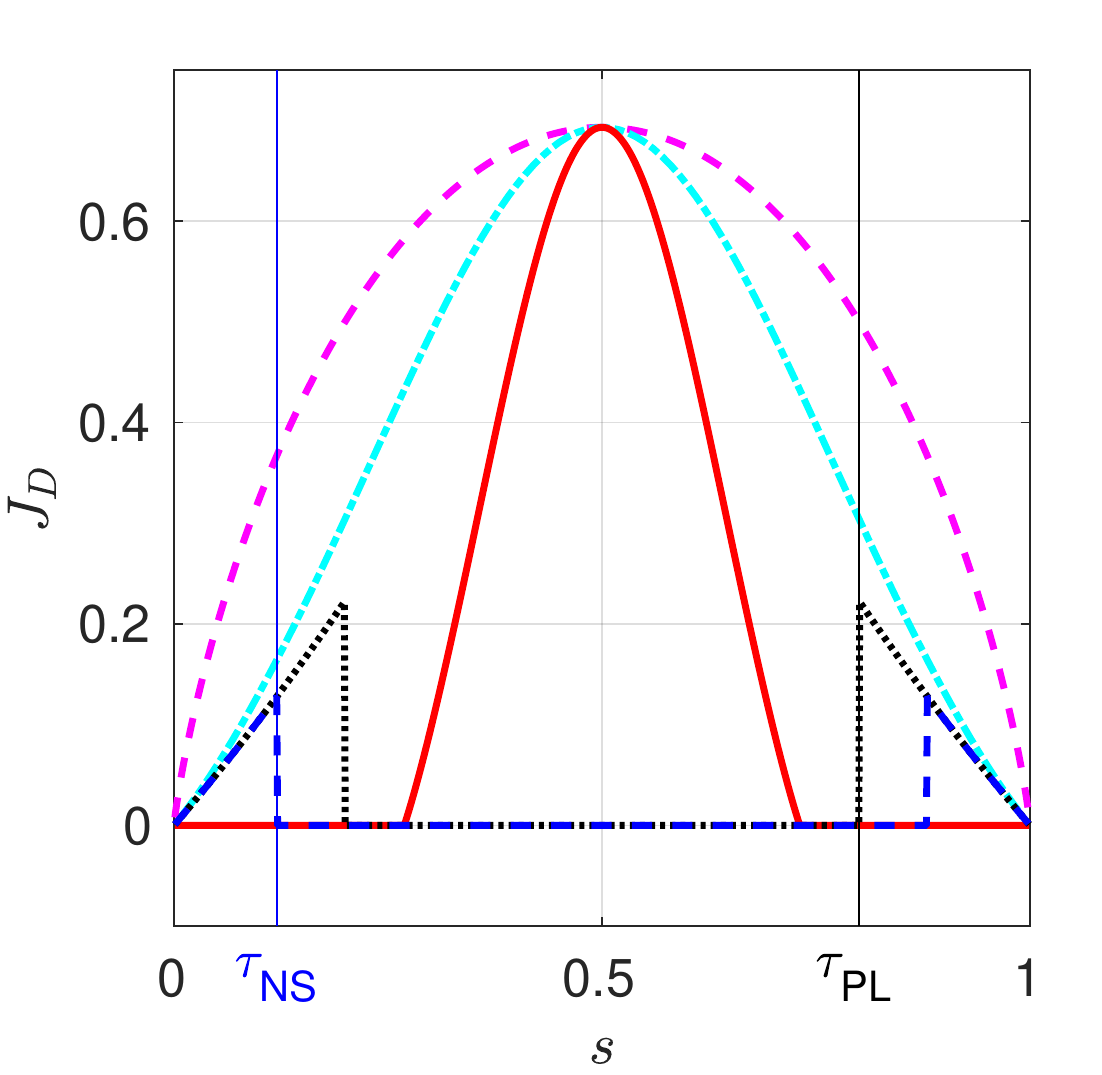}}
\hfill
\subfigure[Distillation Gradient]{\label{fig:c}\includegraphics[width=67mm]{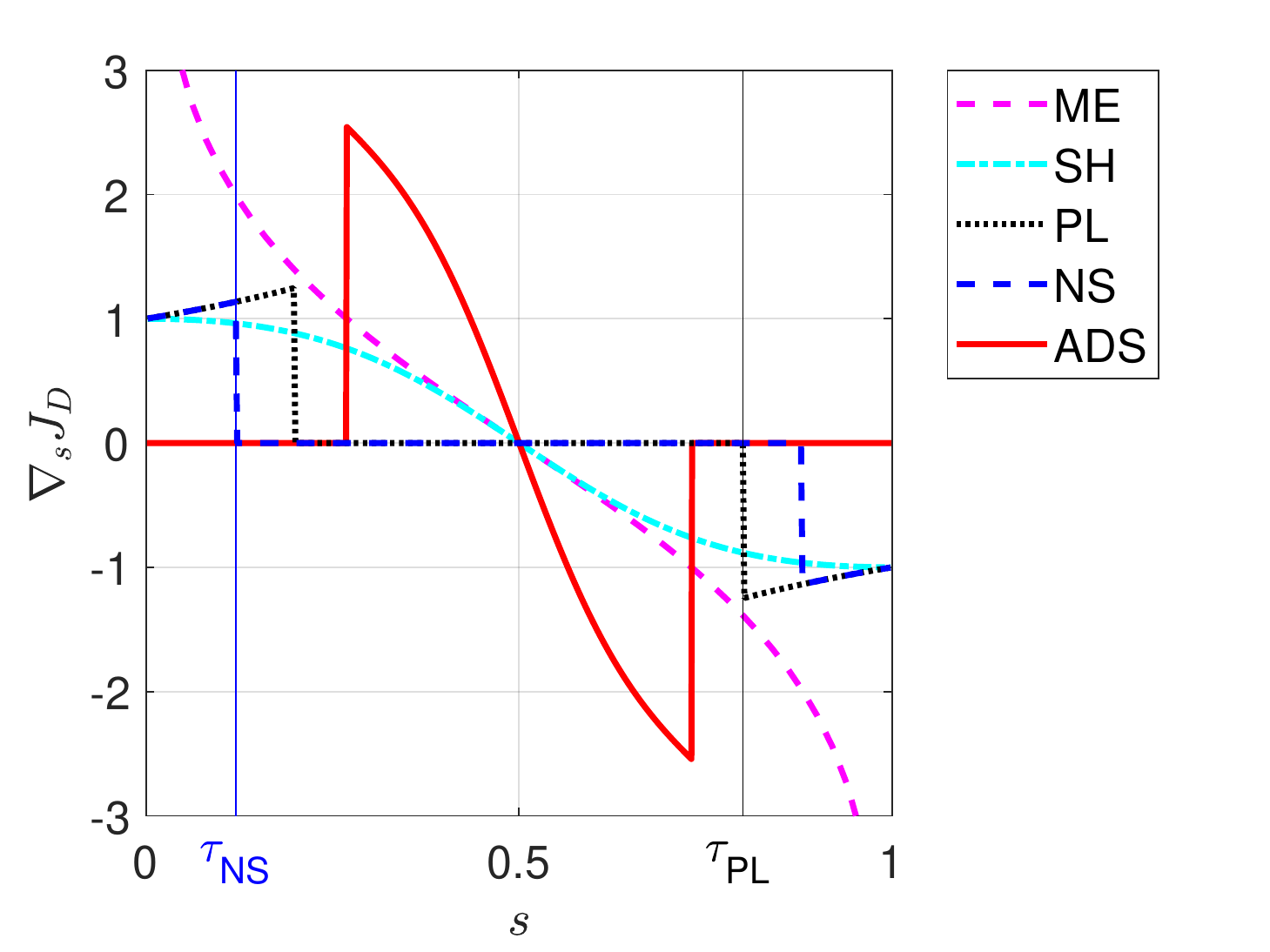}}
\caption{Comparison of different distillation strategies in terms of target probability, distillation loss, and gradient. Regarding SH and \ADS, only the actual losses which take effect are plotted in subfigure (b), where a constant is removed as the target label distribution $\bm{q}$ is stopped gradient. Correspondingly, the distillation gradient of SH and \ADS shown in subfigure (b) is deduced using the actual losses. }
\label{fig:distillation_compare}
\end{figure*}

As supervised loss and consistency regularization are both computed over predictions, unlike other distillation strategies \ADS indeed has an direct impact on these two losses as well. 

In terms of consistency regularization, it restricts each unlabeled sample to reach a consistent prediction with its transformed counterpart. For \ADS, consistency loss $\mathcal{J}_\text{C}$ could be generally formulated as
\begin{equation}\label{eq:consistency}
    \mathcal{J}_\text{C} = \text{Dist}(\text{sparsemax}(\bm{z}), \text{sparsemax}(\bm{z}'))
\end{equation}
where $\bm{z}$ and $\bm{z}'$ are logits vectors of the inspected unlabeled sample and its counterpart, and $\text{Dist}(\cdot)$ could be any proper distance function. Suppose $\text{sparsemax}_i(\bm{z}) = \text{sparsemax}_i(\bm{z}') = 0 $ holds for the $i$-th class. In this case, class $i$ will not contribute to the consistency loss. With more classes like this, the consistency loss of \ADS will focus on a few of confusing classes only. From Eq.~\eqref{eq:softmax}, the condition of equality of $\text{softmax}_i(\bm{z}) = \text{softmax}_i(\bm{z}')$ is much stricter. That means softmax-based consistency will be distracted by the unrelated classes during model training. Taking image classification for an example, an unlabeled image is categorized to \emph{leopard} only if all its transformed counterparts reach a consensus on \emph{leopard}. In practice, the improvement of consistency regularization essentially lies in its constantly confirmed decision among the confusing classes. Intuitively, an augmented \emph{leopard} image might be somehow close to \emph{cat} while it should less similar to \emph{dog}. Thus, putting more attentions on confusing classes are expected to achieve better performance which is also referred in previous research~\cite{gal2016dropout}.

Particularly, we also allow the networks' output of labeled data to pass the sparsemax. However, if the model assigns zero probability to gold label, the entire training sample would be ruled out~\cite{martins2016softmax}. Although there exists some possible workaround like adding a small constant to the probabilities and then do re-normalization, we instead use the following supervised loss as an alternative of cross entropy proposed by the recent work~\cite{DBLP:journals/jmlr/BlondelMN20}  
\begin{equation}
    \mathcal{J}_{\text{S}} = \frac{1}{2}(||\bm{y}-\bm{z}||^2 - ||\text{sparsemax}(\bm{z})-\bm{z}||^2),
\end{equation}
where $\bm{y}$ is the groundtruth encoded by one-hot format. Notice that the $\ell_2$-norm based loss trickily averts the aforementioned optimization dilemma, and the derived gradient w.r.t. $z$ has the closed form, making a friendly backward propagation.



\section{Theoretical Analyses}
In this section, we theoretically justify the principle of \ADS, and then provide the evidence of its superiority from the view of transformation function.  

\subsection{Promoting Informed Predictions}\label{sec:inclined}
We highlight that \ADS does distillation via promoting the informed predictions only, showing an efficient distillation mechanism compared with existing strategies. 

Regarding a two-class case, we denote the neural networks' output as $\bm{z}=(u,0)$. In terms of same $\bm{z}$, suppose $\text{softmax}_1(\bm{z})=s$ and $\text{sparsemax}_1(\bm{z})=s'$. In particular, we have $s\neq s'$ if $u\neq0.5$. Since existing distillation strategies work over softmax output, we consider $s'$ as a function of $s$ and rewrite the distillation loss $\mathcal{J}_{\text{D}}$ (refer to the detailed derivation in Appendix~\ref{appendix:same_pro_space}), making a clear and direct comparison for all methods in the same probability space. Figure~\ref{fig:distillation_compare} plots the target probability, prediction distillation loss, and according gradient versus the first dimensional probability $s$ in terms of each distillation strategy, respectively. We remind readers to read this figure from the shape and trend of every curve other than the scale as they can be equipped with different weights during formulations. Our main conclusions are as follows:

\begin{itemize}
    \item \ADS is the unique strategy focusing on informed predictions only. According to Figure~\ref{fig:distillation_compare}(b), we observe that the proposed \ADS has zero penalty when $s$ is negligible or determinate. Notably, \ADS is viewed as a corrective SH by masking out the determinate predictions to avoid introducing overconfident risk.  
    \item PL and NS are consistent strategies for binary classification case. PL and NS have the complementary philosophy from Figure \ref{fig:distillation_compare}(a). Given proper thresholds, they are shown to have the same form of loss and gradient according to Figure~\ref{fig:distillation_compare}(b) and (c).
    \item ME has the unbounded gradients. ME takes the highest loss value but produces zero gradient on ambiguous predictions according to Figure~\ref{fig:distillation_compare}(b) and (c). Meanwhile, when $s$ is relatively small or large, the gradient of ME is large, exposed to the risk of unstable optimization. 
\end{itemize}

From above analyses, \ADS will outperform other distillation strategies if the assumption holds that leveraging the informed predictions only are sufficient for distillation. From Figure~\ref{fig:distillation_compare} we realize the terminology ``informed" is factually specified by a corresponding threshold in the probability space of softmax output. Now we are exploring a more general case where data 
could be categorized to one of the multiple classes.

\begin{theorem}\label{theorem1}
For an unlabeled sample $\bm{u}$ whose softmax output is $\bm{p} \in \mathbb{R}^K (K>2)$, the distillation loss $\mathcal{J}_\text{D}\equiv0$ for \ADS in Eq.~\eqref{eq:J_d} holds if $p_{(1)}> ep_{(2)}$, where $p_{(1)}, p_{(2)}$ are the first two largest coordinate of $\bm{p}$ and $e$ is Euler number.
\end{theorem}
\begin{proof}
Let $\bm{z}\in \mathbb{R}^K$ denote the logits of $\bm{u}$, i.e., $\bm{z}=f(\bm{u};\bm{\theta})$, and $\bm{p}=\text{softmax}(\bm{z})$. 
According to the definition of softmax, i.e., Eq.~\eqref{eq:softmax}, we have $p_i = \frac{e^{z_i}}{C}$ for $i=1,2,...,K$, where $C=\sum_ke^{z_i}$. Then we can rewrite $z_i = \ln(Cp_i)$, for $i=1,2,...,K$. 
Let $p_{(1)}, p_{(2)}$ are the first two largest coordinates of $\bm{p}$, and we have element expression
\begin{equation}\label{eq:theorem1}
z_{(1)}=\ln(Cp_{(1)}), \quad z_{(2)}=\ln(Cp_{(2)}),
\end{equation}
where $z_{(1)}, z_{(2)}$ denote the corresponding first two largest coordinates of $\bm{z}$. 
Apart from the trivial case of ${p}_{(1)}= {p}_{(2)}=...={p}_{(K)} = \frac{1}{K}$,
$\mathcal{J}_\text{D}= 0$ holds iff $\text{sparsemax}(\bm{z})$ reaches one-hot. According to the closed-form solution of sparsemax in Section \ref{sec3.1}, we obtain
\begin{equation}\label{eq:theorem2}
    1+2z_{(2)} \le z_{(1)}+z_{(2)}.
\end{equation}
Taking Eq. \eqref{eq:theorem1} into Eq. \eqref{eq:theorem2}, we arrive
\begin{equation} \label{eq:theorem3}
\begin{split}
1+\ln(Cp_{(2)})\le \ln(Cp_{(1)}) \Rightarrow p_{(1)} \ge ep_{(2)},
\end{split}
\end{equation}
which completes the proof.
\end{proof}
From Theorem~\ref{theorem1} we can see that the ``informed" predictions spoken of in the output space of softmax refer to the categorical probability $\bm{p}$ which satisfies $p_{(1)} < ep_{(2)}$ and $\bm{p}\neq \frac{1}{K}\bm{1}$. \textbf{That means unlike existing distillation strategies, \ADS does not encourage the relatively certain predictions to further become extremely certain since the distillation loss $\mathcal{J}_\text{D}\equiv0$ holds as long as $p_{(1)}\ge ep_{(2)}$, and thus the issue of overconfident predictions are mitigated.} The following corollary quantify the soft threshold used by \ADS. 

\begin{corollary}
For a $K$-way semi-supervised classification problem, the determinate predictions and negligible predictions for \ADS are masked out by the sample dependent threshold $\theta_1 \in [ \frac{e}{e+K-1},\frac{e}{e+1}]$ and $\theta_2 \in [ \frac{e^{\rho}}{\rho+e^{\rho}(K-\rho)},\frac{e^{\rho}}{\rho+e^{\rho}}]$ in the corresponding softmax output space, respectively, where $e$ is Euler number and $\rho$ is the population of non-zero predictions.
\end{corollary}
We leave the proof of this corollary to Appendix~\ref{proof_corollary}. Notably, we clarify that the determinate predictions are spoken of in the context of multi-class classification where only a single class is the groundtruth. In other words, a prediction is said determinate in terms of its potential to be the real label. In addition, the negligible predictions are handled in a similar manner, but its population for a single unlabelled sample is at least 1. In particular, for \ADS the negligible predictions do not coexist with the determinate prediction.

\subsection{Sparsemax Facilitating Entropy Minimization}\label{sec:EM}
Entropy minimization is a golden principle that has been demonstrated effective in existing SSL research \cite{grandvalet2005semi,miyato2018virtual,oliver2018realistic}. 
In this section, we point out this principle is fundamentally related to the probability transformation function used in neural networks, e.g., softmax activation.
It typically involves projecting a logits vector $\bm{z}$ on the probability simplex with an optimized problem of
\begin{equation} \label{eq:general_obj}
     \bm{p^\star} = \argmin_{p \in \Delta^{K-1}} \{ -\langle\bm{z},\bm{p}\rangle - H(\bm{p}) \},
\end{equation}
where $\Delta^{K-1}$ is the probability simplex with freedom of $K-1$, and $-H(\bm{p})$ is a convex function, serving as a regularizer. 

When $H(\bm{p})$ is implmented using Shannon entropy, i.e., $H(\bm{p})=-\sum_k p_i \log p_i$, the closed-form solution of Eq.~\eqref{eq:general_obj} is the softmax transformation, a.k.a. maximum entropy transformation. When $H(\bm{p})$ is replaced with Gini entropy, namely $H(\bm{p})= \frac{1}{2}\sum_k p_i (1-p_i)$, the solution of Eq.~\eqref{eq:general_obj} is equivalent to the sparsemax transformation, i.e., Eq~\eqref{eq:sparsemax}. 

Back to the minimum entropy principle, existing distillation methods built on softmax can be viewed as minimizing the prediction entropy in a post-processing manner \cite{DBLP:conf/nips/ChenKSNH20}, which is in conflict with the function of the regularizer $H(\bm{p})$ in Eq.~\eqref{eq:general_obj}.
As Shannon entropy is a stronger penalty than Gini entropy, softmax would intensify contradictions compared with sparsemax~\cite{DBLP:journals/jmlr/BlondelMN20}. 
Therefore, we argue that our \ADS facilitates entropy minimization since it reaches a better trade off between probability transformation  Eq.~\eqref{eq:general_obj} and predication distillation~$\mathcal{J}_\text{D}$. 

\section{Experiments}

\subsection{Experimental Setup}
We conduct empirical comparisons following the experimental setting of \cite{chen2020negative} except FixMatch for which we adopt the codebase from \cite{sohn2020fixmatch}. For fair comparison, the default network architecture for all datases except MNIST is Wide ResNet-28-2~\cite{zagoruyko2016wide} with 1.5M parameters. Regarding MNIST, we employ a 7-layer convolutional neural network. The batch size for unlabeled data is 64. In terms of the labeled data, the batch size is set as the number of labeled data if it is smaller than 64, and set as 64 otherwise. We do not exhaustively adjust the network parameters for different benchmarks, such as the number of scales and filters, in order to best reproduce the results. The mean and variance of five independent running under different random seeds are reported for fair~comparison. Throughout all the experiments involving \ADS, we use $r=2$ as a default setting as we observe this value achives stable performance for different datasets. 

In particular, we use ``X+Y" to dub a method by adding a distillation strategy Y to a SSL algorithm X, and use ``X-Y" to dub a SSL algorithm X whose distillation component is replaced by Y. 

\begin{figure*}[t]
\centering     
\subfigure[ME]{\label{lossfig:a}\includegraphics[width=34mm]{  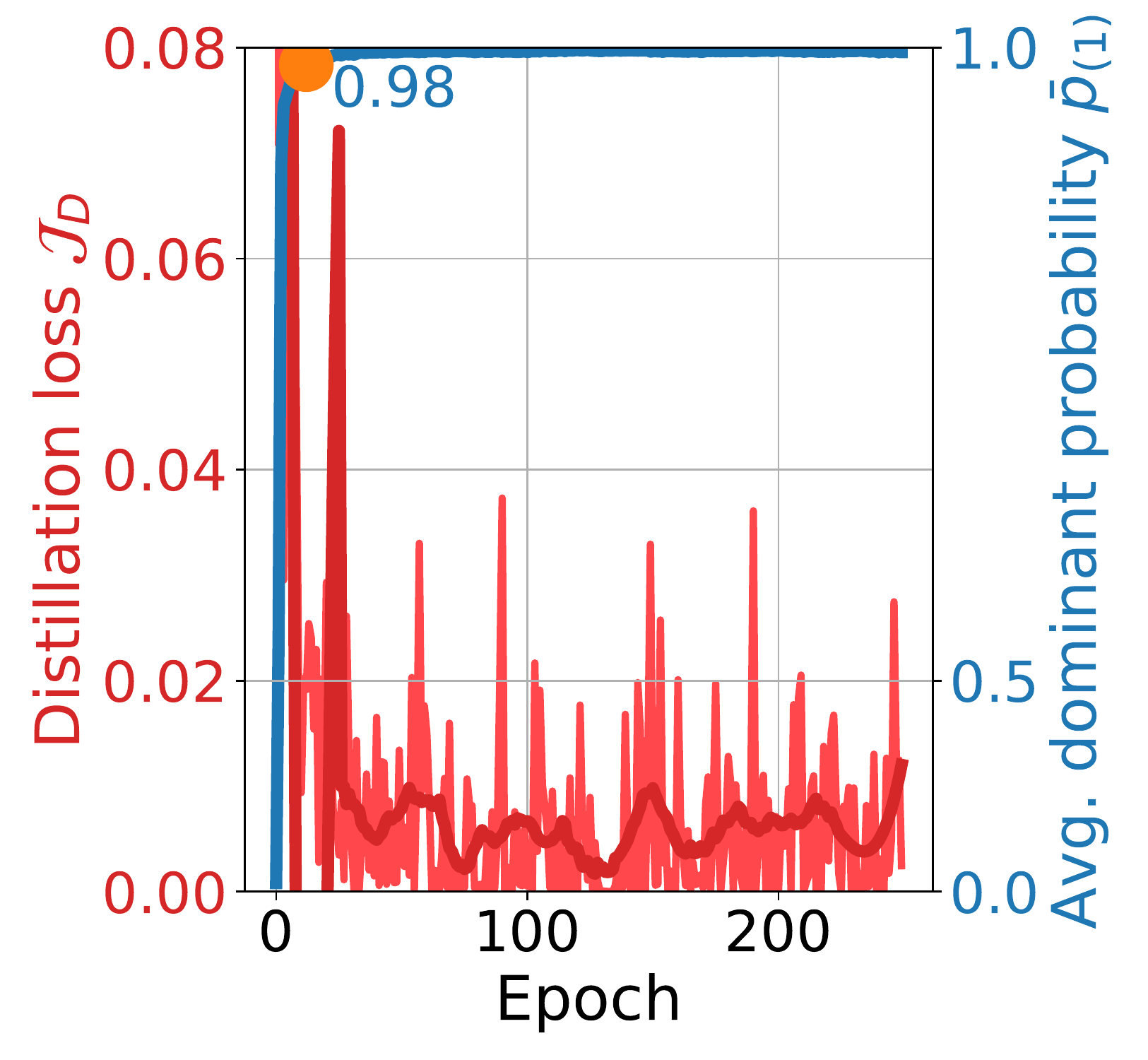}}
\hspace{0em}
\subfigure[SH]{\label{lossfig:b}\includegraphics[width=34mm]{  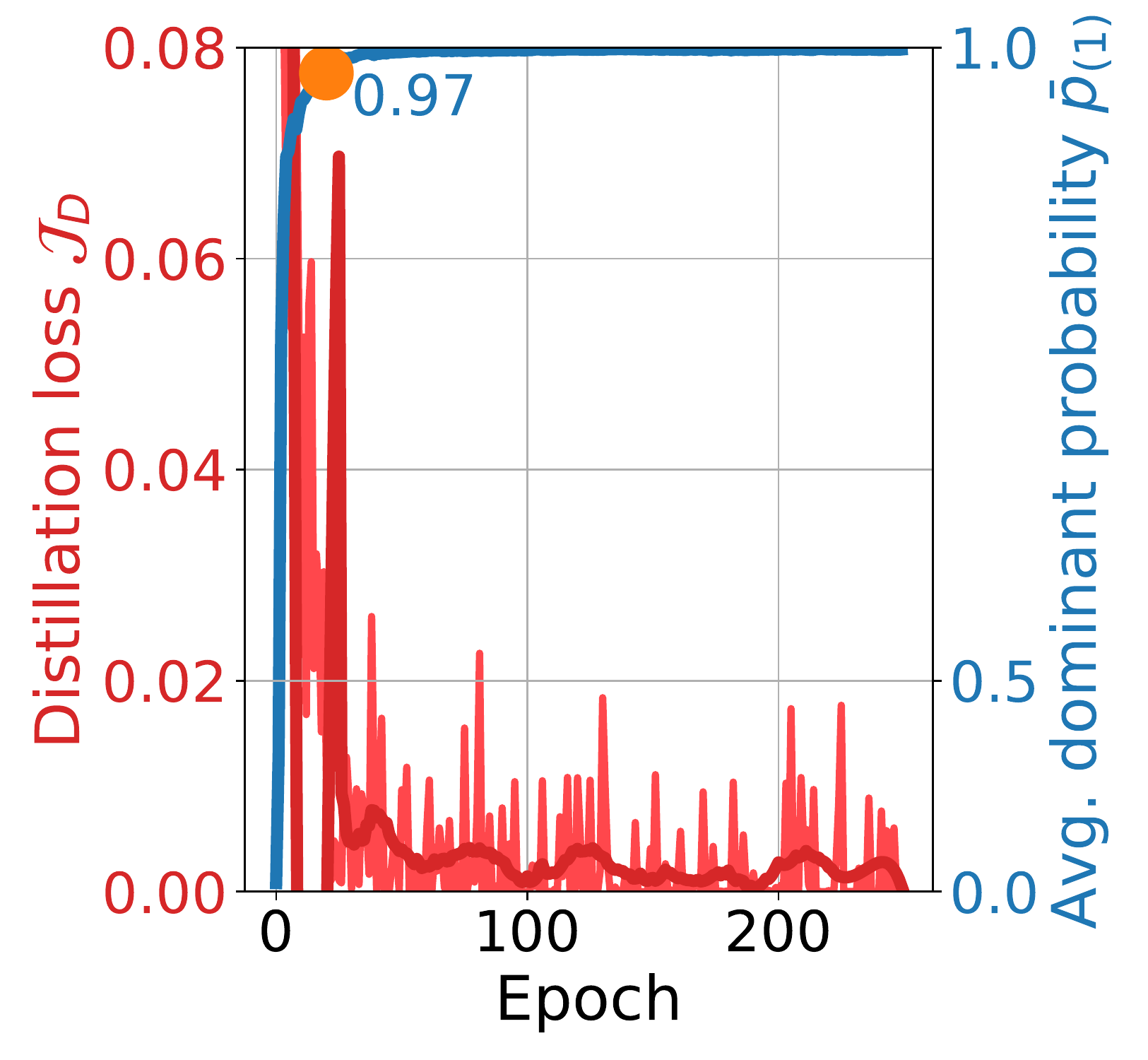}}
\hspace{0em}
\subfigure[PL]{\label{lossfig:c}\includegraphics[width=34mm]{  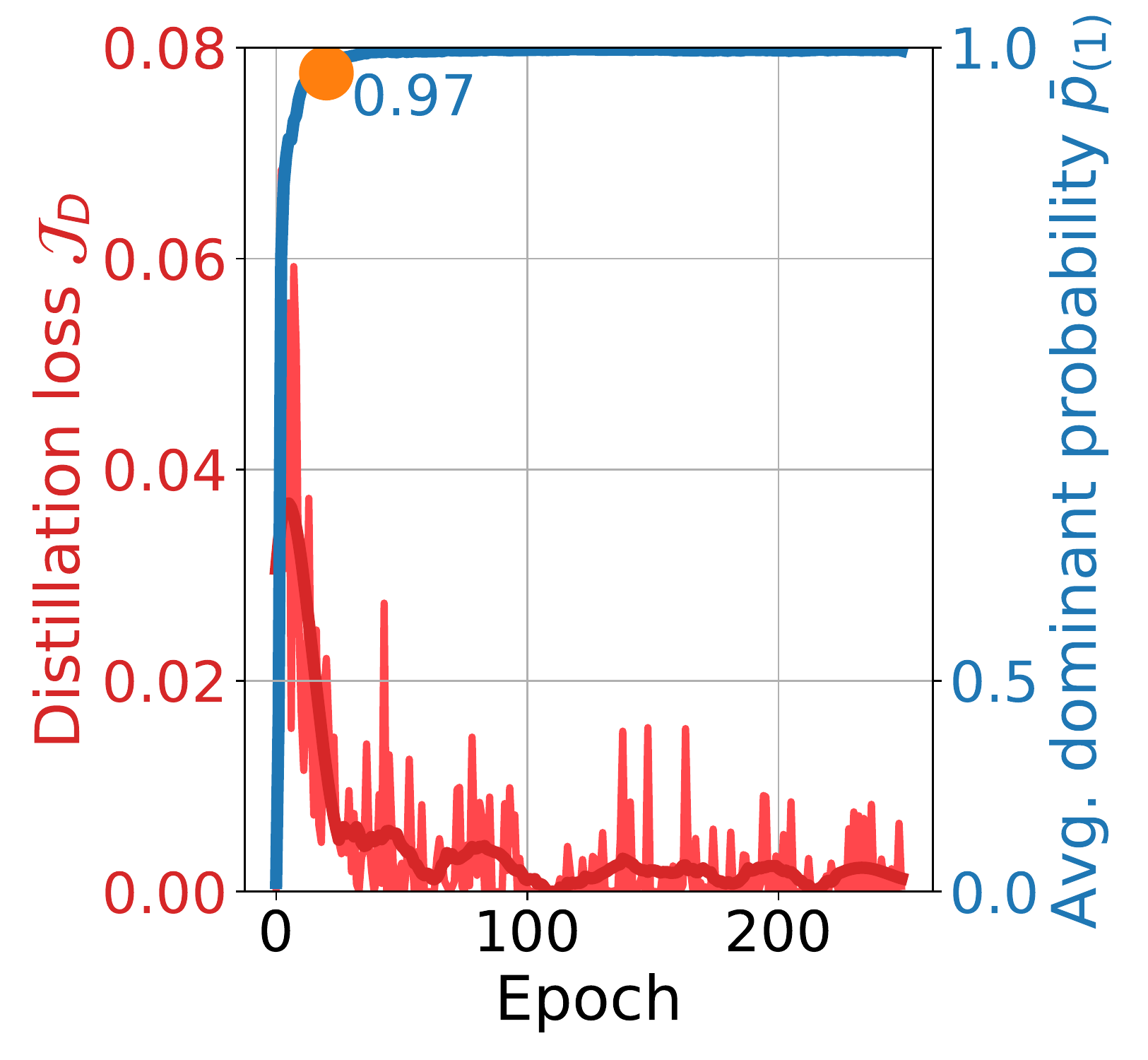}}
\hspace{0em}
\subfigure[NS]{\label{lossfig:d}\includegraphics[width=34mm]{  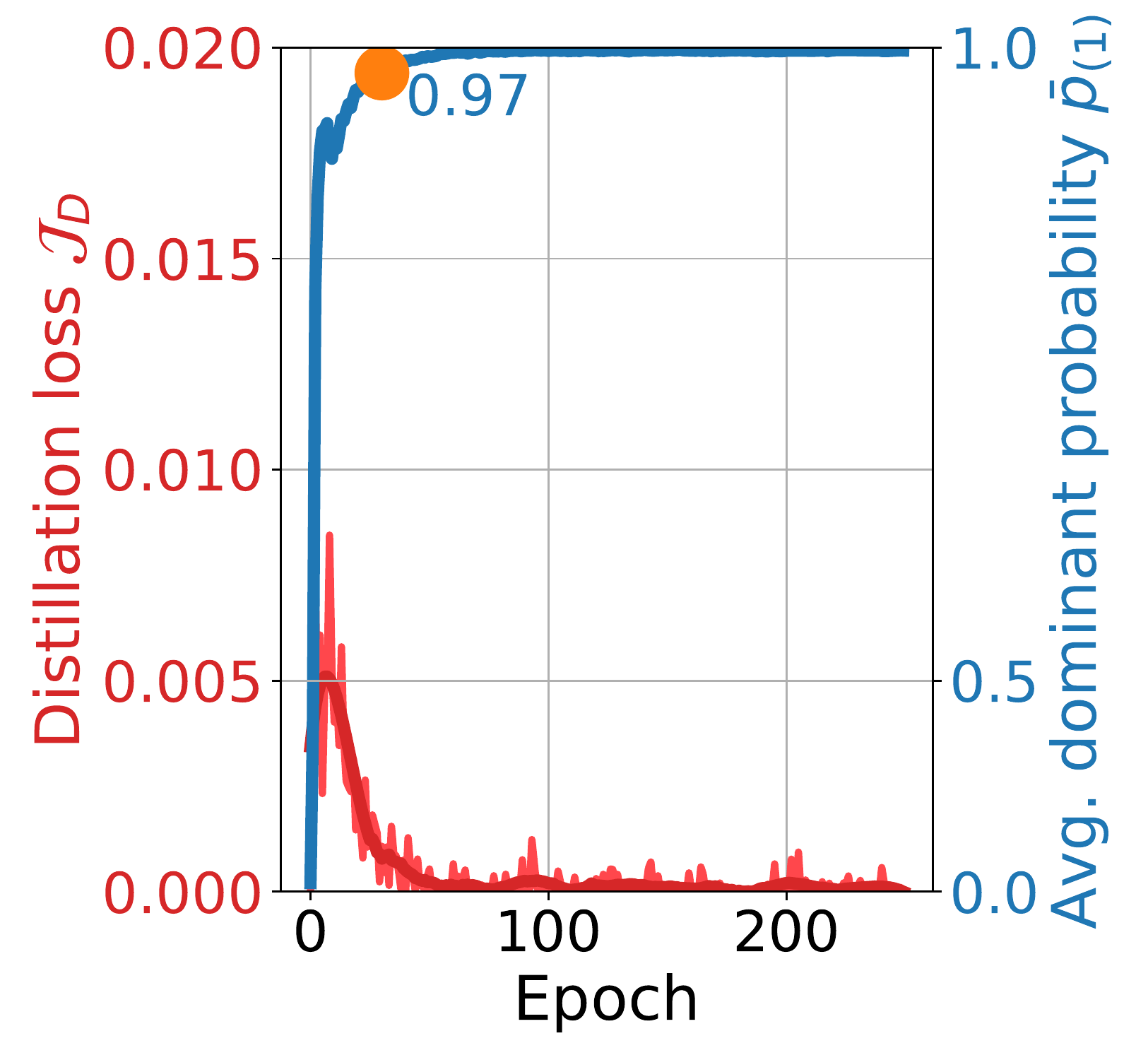}}
\hspace{0em}
\subfigure[\ADS]{\label{lossfig:e}\includegraphics[width=34mm]{  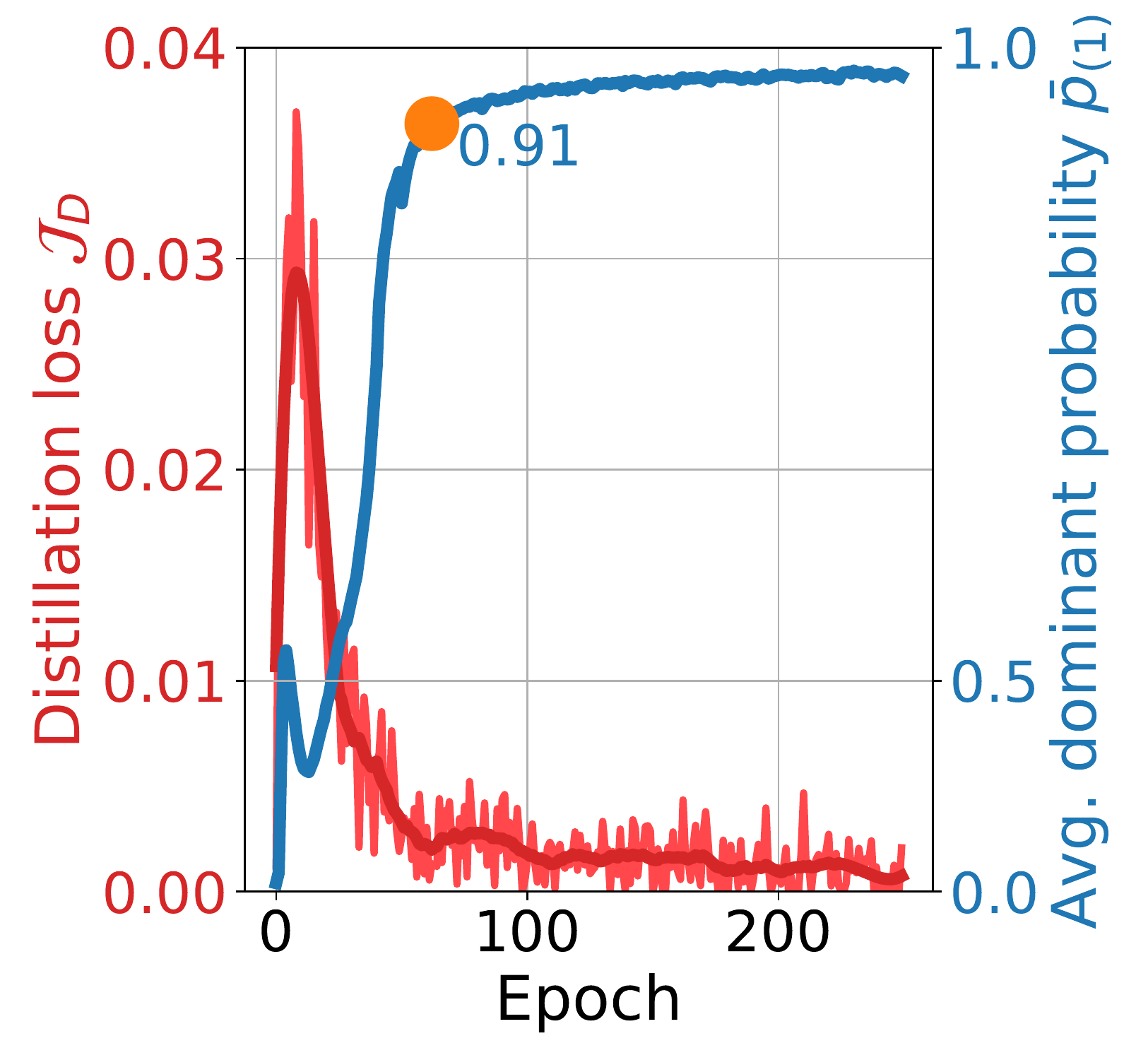}}
\caption{\label{fig:loss_pro}Converged curves of distillation loss $\mathcal{J}_{\text{D}}$ and average dominant probability $\bar{p}_{(1)}$ for unlabeled training samples on MNIST datset. We smooth the loss values for a better visualization. For \ADS, $\bar{p}_{(1)}$ is collected and calculated by replacing sparsemax with softmax which does not change the training process. }
\end{figure*}


\subsection{Study on VAT}\label{sec:5.2}
\begin{table}[!t]
\centering
\caption{\label{table:VAT} Test error (\%) of various distillation strategies based on VAT. The best results are marked in bold.}
\begin{tabular}{ccc}
\toprule[1.3pt]
Methods  & MNIST (20 labels) & CIFAR-10 (4,000 labels)\\
\midrule
VAT & $23.76\pm1.18$ & $14.72 \pm 0.23$\\
VAT+ME& $20.64 \pm 1.28$ & $14.34 \pm 0.18$\\
VAT+SH & $18.45\pm 1.09$ & $12.90 \pm 0.25$\\
VAT+PL &$19.72.\pm 1.35$ & $14.15 \pm 0.14$ \\
VAT+NS & $19.36\pm 1.17$ &  $13.94 \pm 0.10$\\
VAT+\ADS&$\bm{14.52 \pm 1.03}$ & $\bm{12.40 \pm 0.31}$\\
\bottomrule[1.3pt]
\end{tabular}
\end{table}
Virtual Adversarial Training~\cite{miyato2015distributional} (VAT) serves as a powerful SSL algorithm without requiring manual data augmentation~\cite{oliver2018realistic}. It contains a consistency loss which has a similar form with Eq.~\eqref{eq:consistency}, but optimizes the adversarial perturbation $r_{\text{adv}}=\epsilon \frac{g}{||g||}$, where $\epsilon$ is a predefined perturbation scale and $g$ is approximated by the gradient on a randomly sampled unit vector $r$, that is
\begin{equation}
    \nabla_r \text{Dist} (\text{softmax}(f(x;\bm{\theta})), \text{softmax}(f(x+r;\bm{\theta}))).
\end{equation}
VAT is suggested to be improved if it is added to a minimum entropy according to~\cite{miyato2018virtual}. Based on this fact, we implement various distillation strategies on VAT for a direct and fair comparison. To this end, we construct each baseline by replacing the distillation loss $\mathcal{J}_{\text{D}}$ of Eq.~\eqref{eq:J_total} with differential distillation strategies. Experiments are conducted on MNIST and CIFAR-10 with 20 and 4,000 labeled training examples, respectively.

Table \ref{table:VAT} presents the test error of various distillation strategies based on VAT. It is observed that each distillation loss assists semi-supervised learning and obtains the better performance. More importantly, our \ADS achieves the lowest test error compared with both vanilla VAT and other variants. In particular, the improvement to the second-best on two datasets is around $4.7\%$ and $1.5\%$, respectively. 

To have a close look at the distillation component for MNIST dataset on which a better improvement has been achieved. For each method we visualize the distillation loss $\mathcal{J}_{\text{D}}$ and average dominant probability. Average dominant probability is calculated over all unlabeled training samples  \begin{small}$\bar{p}_{(1)}=\frac{1}{|\mathcal{U}|}\sum_{\bm{x}\in \mathcal{U}}p_{(1)}$\end{small} which reflects the certainty of unlabeled predictions.
We record two measures at each epoch and results are shown as Figure~\ref{fig:loss_pro}.

In terms of our \ADS, the distillation loss decreases stably with the increase of average dominant probability. The curve of $\bar{p}_{(1)}$ reaches a cusp and levels off, with more modest increases after $\bar{p}_{(1)}=0.91$. For the other four strategies, the distillation loss decreases dramatically, and the average dominant probability surges within a small number of epochs. 
Clearly, their values of $p_{(1)}$ locate at the cusp, which are all much larger than that of \ADS’s. That is to say, they aggressively optimize the learning model to an extremely certain state but inevitably introduce overconfident distillations. Instead, our \ADS gradually optimizes the model by using the informed predictions, averting the overconfident risks naturally. Let $\mathcal{U}_+$ and $\mathcal{U}_-$ denote the correctly and incorrectly classified unlabeled 
training examples, and we can rewrite average dominant probability as \begin{small}$\bar{p}_{(1)}=\frac{1}{|\mathcal{U}_1|}(\sum_{\bm{x}\in \mathcal{U}_+}p_{(1)} + \sum_{\bm{x}\in \mathcal{U}_-}p_{(1)} )$\end{small}. For the correct ones, they are expected to be more certain, but are not supposed to be extremely certain in practice. An evidence for this claim is that in supervised task, penalizing the low-entropy prediction has been demonstrated to be beneficial to model generalization. One of the simplest trick is label smoothing~\cite{pereyra2017regularizing}. For the incorrect ones, their dominant prediction should be small as they provide the wrong optimization direction. To sum up, the relative smaller $\bar{p}_{(1)}$ than competitors suggests better performance by considering two aspects. This is another hindsight by reviewing existing methods.

\begin{table*}[!htbp]
\centering
\caption{\label{table_mixmatch_fixmatch}Performance comparison on four benchmarks. The best performance are marked as bold in two separate blocks.}
\begin{tabular}{cccccccc}
\toprule[1.3pt]
\multirow{2}{*}{Methods}  &\multicolumn{2}{c}{CIFAR-10}  & \multicolumn{2}{c}{CIFAR-100} & \multicolumn{2}{c}{SVHN} & STL-10 \\ &250 labels & 4,000 labels  & 2,500 labels & 10,000 labels & 250 labels & 1,000 labels & 1,000 labels \\
\midrule
MixMatch  &$14.49 \pm 1.60$ &$7.05 \pm 0.10$ &$39.94 \pm 0.37$ &$33.72\pm 0.33$ &$3.75 \pm 0.09$ & $3.28 \pm 0.11$ & $22.20 \pm 0.89$\\
MixMacth+NS   & $12.48 \pm1.21$              &$6.92 \pm 0.12$ & $39.74\pm 0.21$&$33.45\pm 0.19$ &$3.38 \pm 0.08$ & $3.14 \pm 0.11$ &$21.74 \pm 0.33$\\
MixMatch-ADS &$ \bm{10.87 \pm 0.92}$  &$\bm{6.46 \pm 0.21}$ &$\bm{39.10\pm 0.25}$&$\bm{33.18 \pm 0.54}$  &$\bm{2.63\pm 0.06}$&$\bm{2.42 \pm 0.14}$ &$\bm{19.20 \pm 0.31}$\\
\midrule
FixMatch  &$\bm{6.81 \pm 0.42}$ & $5.92 \pm 0.14$ &$37.42 \pm 0.38$ &$29.57 \pm 0.11$ &$2.64 \pm 0.64$ &$2.36\pm0.19$ & $11.12\pm 0.63$\\
FixMatch-ADS & $7.18 \pm 0.37$    &$\bm{5.21 \pm 0.18}$ & $\bm{37.28 \pm 0.42}$& $\bm{27.96 \pm0.13}$ &$\bm{2.15 \pm 0.45}$ & $\bm{2.07 \pm 0.15}$& $\bm{10.43 \pm 0.47}$\\
\bottomrule[1.3pt]
\end{tabular}
\end{table*}

\subsection{Improvement on Advanced SSL Algorithms}\label{sec:5.3}
To evaluate the efficacy of our \ADS on the SSL algorithm which benefits from data augmentations, we plug \ADS in two state-of-the-art models MixMatch \cite{berthelot2019mixmatch} and FixMatch \cite{sohn2020fixmatch}. Since vanilla MixMactch has already incorporated SH while vanilla FixMatch has adopted PL as their distillation methods separately, we do not conduct the corresponding ablation study. Notably, ME does not construct an explicit target label, and thus cannot be applied to data augmentations based SSL algorithms, which requires distilling explicit labels for augmented unlabeled data.  
Following~\cite{chen2020negative}, we evaluate the remaining baselines on four benchmarks CIFAR-10, CIFAR-100, SVHN, and STL-10 and present the experimental results in Table~\ref{table_mixmatch_fixmatch}.



Table~\ref{table_mixmatch_fixmatch} shows that \ADS based methods achieve significant improvement over vanilla Mixmatch, Fixmatch, and their variants, demonstrating the superiority of \ADS over other distillation strategies, i.e., SH, PL, NS. In addition, we observe that MixMatch+NS also achieves some improvement over vanilla Mixmatch. Interestingly, this method can be viewed as a simple compensation of overusing the extremely certain predictions (See the distillation loss of SH and NS in Figure~\ref{fig:distillation_compare}(b)). From this aspect, \ADS alone is functionally similar to this combination and MixMatch-ADS is shown to outperform the two baselines. FixMatch-ADS most time improves FixMatch except for CIFAR-10 with 250 labels. This failure case may result from the occasion that PL can deny the strong augmentations that are out of data distribution once its predictions on weak augmentation are not very satisfactory. In a ward, all these observations are consistent with our claim that promoting informed predictions are beneficial for SSL model training to distill confident and reliable output.





\subsection{Safety with Different Backbone Structures}\label{sec:safety}
\begin{figure}[t]
\centering     
\subfigure[Sparsity]{\label{candfig:a}\includegraphics[width=40mm]{  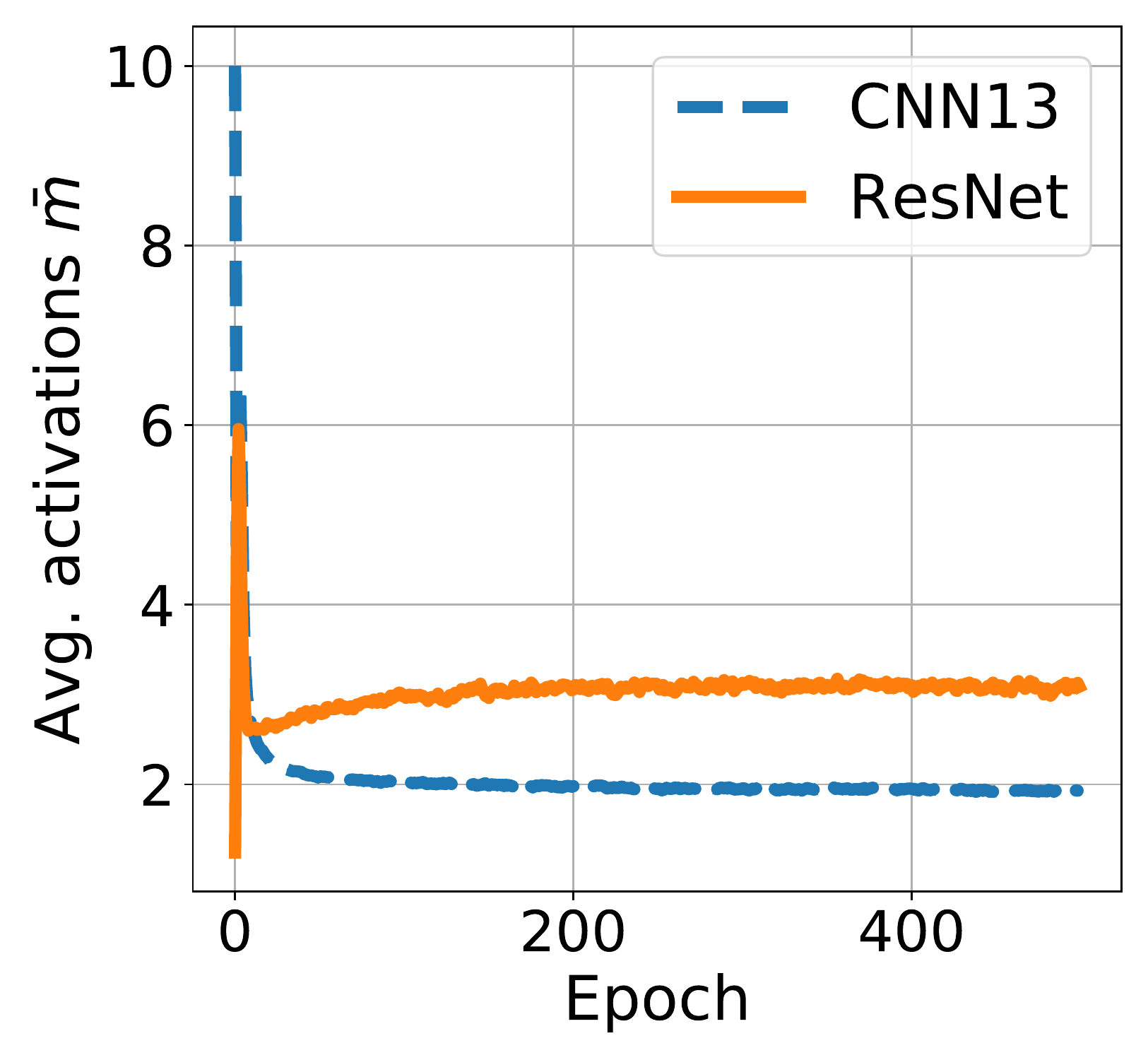}}
\subfigure[Top-$m$ accuracy]{\label{candfig:b}\includegraphics[width=41mm]{  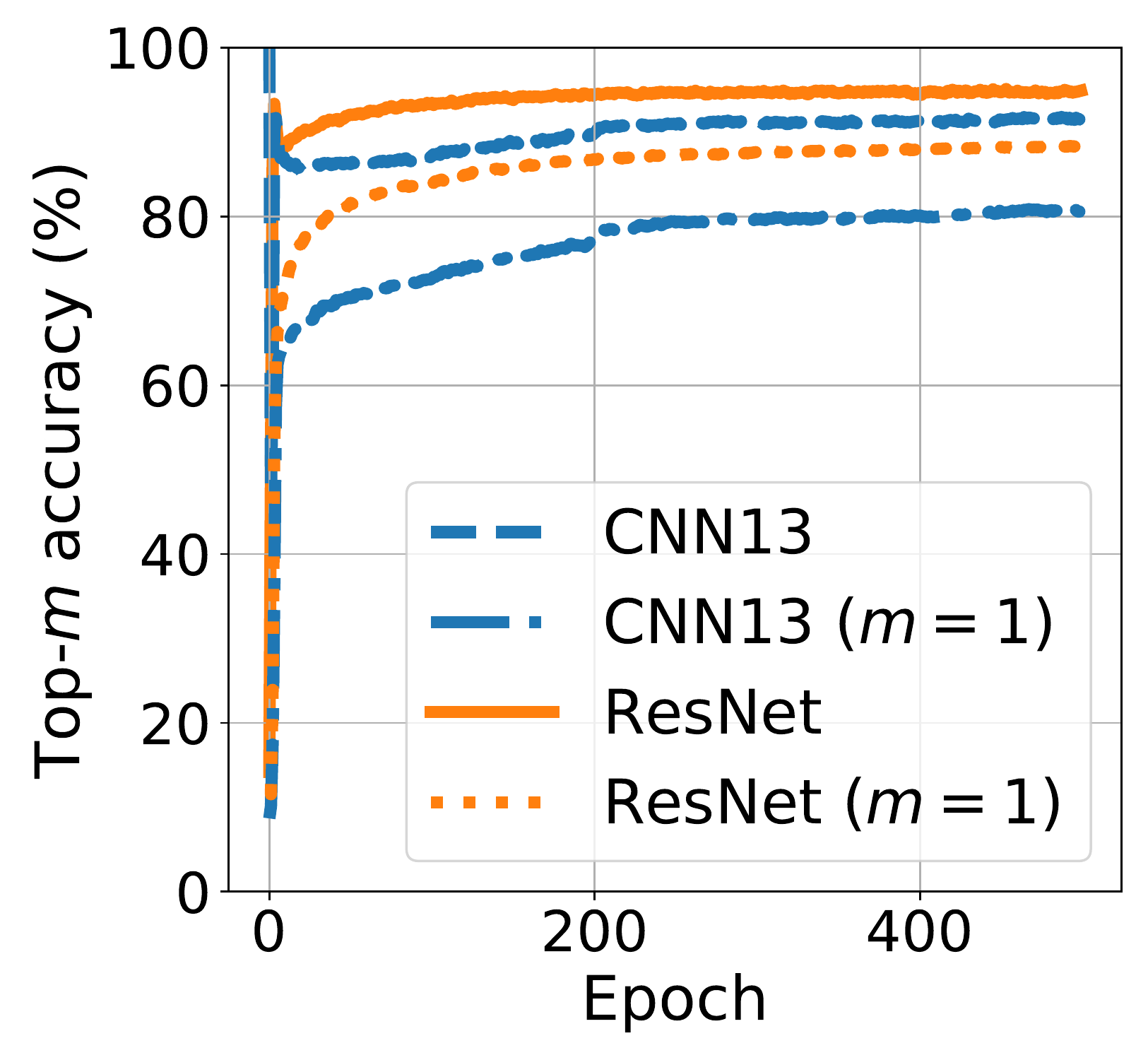}}
\subfigure[Top-$m$ selection]{\label{candfig:c}\includegraphics[width=41mm]{  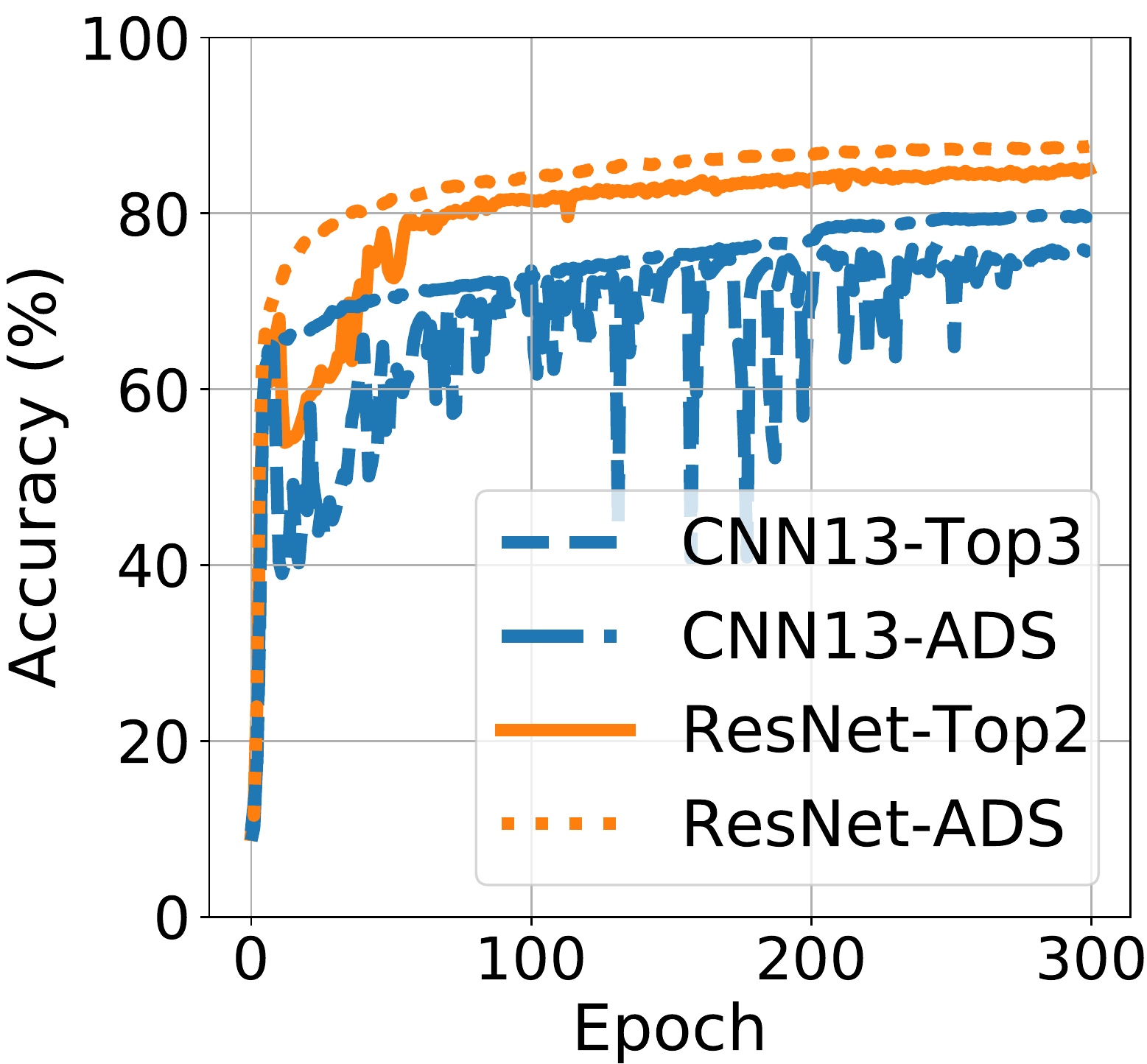}}
\caption{\label{fig:property}The safety study of candidates selection in terms of two backbones ResNet and CNN13. (a) The average sparse activations $\bar{m}$ on unlabeled training data. (b) Top-$m$ accuracy comparison where $m$ is example-wise sparsity. (c) Standard accuracy comparison with globally fixed Top-$m$ selection.}
\end{figure}
Intuitively, it should be safe to do candidates selection when every class is initialized to be evenly activated, while a concern arises if the neural network is biasedly initialized to prefer some specific classes which fail to cover the true label. To address this concern, we apply two different neural network architectures on CIFAR-10 with 250 training labels as an example. Apart from the default ResNet which is initialized with the normal distribution, CNN13~\cite{tarvainen2017mean} is employed which is typically initialized with the uniform distribution. We record the average activations $\bar{m}$ and Top-$m$ accuracy over all unlabeled training data in Figure~\ref{fig:property}, where example-wise $m$ is the number of non-zeros entries. Note that ``Top-$m$ accuray" denotes the success rate of partial activations including the true label, and it degenerates to the standard accuracy if $m=1$. 


Figure~\ref{fig:property}(a) shows that CNN13 activates all the classes at the beginning ($\bar{m}=10$) and then rules out unrelated classes till convergence. Contrastingly, the number of activations for ResNet surges from the one-hot state ($\bar{m}=1$) in the first few of epochs, then drops to a low value, gradually reaches to a steady value afterwards. Figure~\ref{fig:property} (b) shows that ResNet achieves better performance than CNN13 in terms of both two measurements although it starts with a biased initialization (Figure~\ref{fig:property} (a)), experimentally demonstrating that initialization is not an issue for candidates selection of \ADS. In addition, since standard accuracy is upper bounded by Top-$m$ accuracy as we expect, there exists a considerable gap of around 5\% for ResNet and 10\% for CNN13. This observation suggests that the selected candidates of \ADS are often meaningful.

Notably, neither ResNet or CNN13 converges to one-hot format in the above experiment; the converged average activations $\bar{m}$ of them are around 3 and 2, respectively. A possible reason for this phenomenon is that the distillation loss of MixMatch is explicitly defined on the augmented unlabeled data despite their qualities instead of the natural ones, and our model $\emph{rejects}$ to admit any class as the true label for during distillation. A further question is raised that if they are equivalent to the counterparts in which only Top-$3$ and Top-$2$ predictions are picked (with the remaining truncated) and distilled during the entire training stage. This thought is also known as set-valued classification~\cite{chzhen2021set}, and we compare them as a new group of experiment. 

Figure~\ref{fig:property}(c) presents the accuracy comparison between \ADS and fixed Top-$m$ selection. We can see both two backbones have the better accuracy when \ADS is used, where their activations are dynamically changed indicated by Figure~\ref{fig:property}(a). In addition, it is also observed that Top-$m$ selection suffers the intense fluctuation on accuracy, which is rooted in the unstable optimization of the plain truncation. Particularly, the accuracy drop for CNN13 is severer than ResNet, which suggests another benefit of residual structure.

\subsection{Observation of Prediction Histograms}
\begin{figure*}[t]
\centering     
\subfigure[epoch=1]{\label{epochfig:a}\includegraphics[width=24mm]{  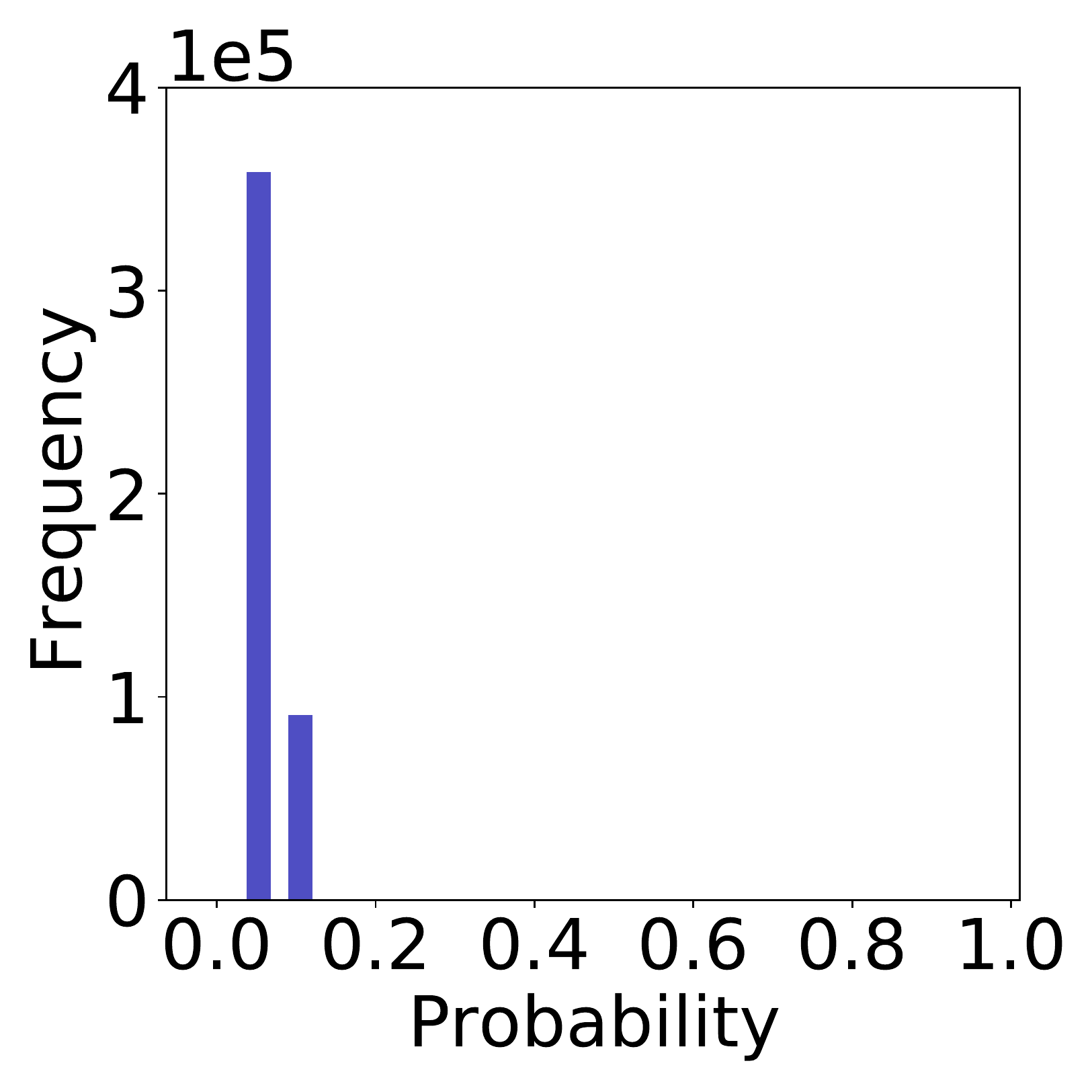}}
\hspace{0em}
\subfigure[epoch=10]{\label{epochfig:b}\includegraphics[width=24mm]{  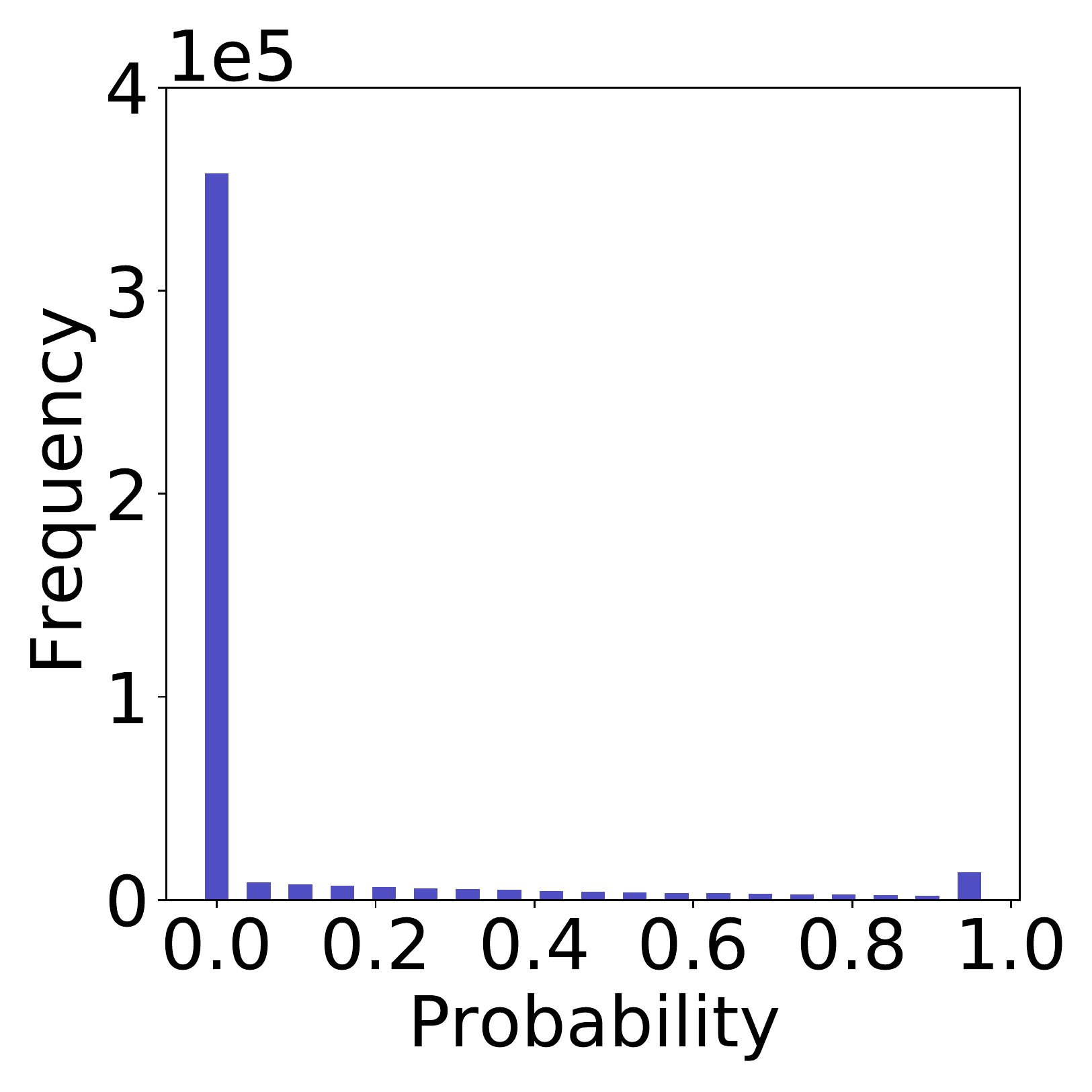}}
\hspace{0em}
\subfigure[epoch=20]{\label{epochfig:c}\includegraphics[width=24mm]{  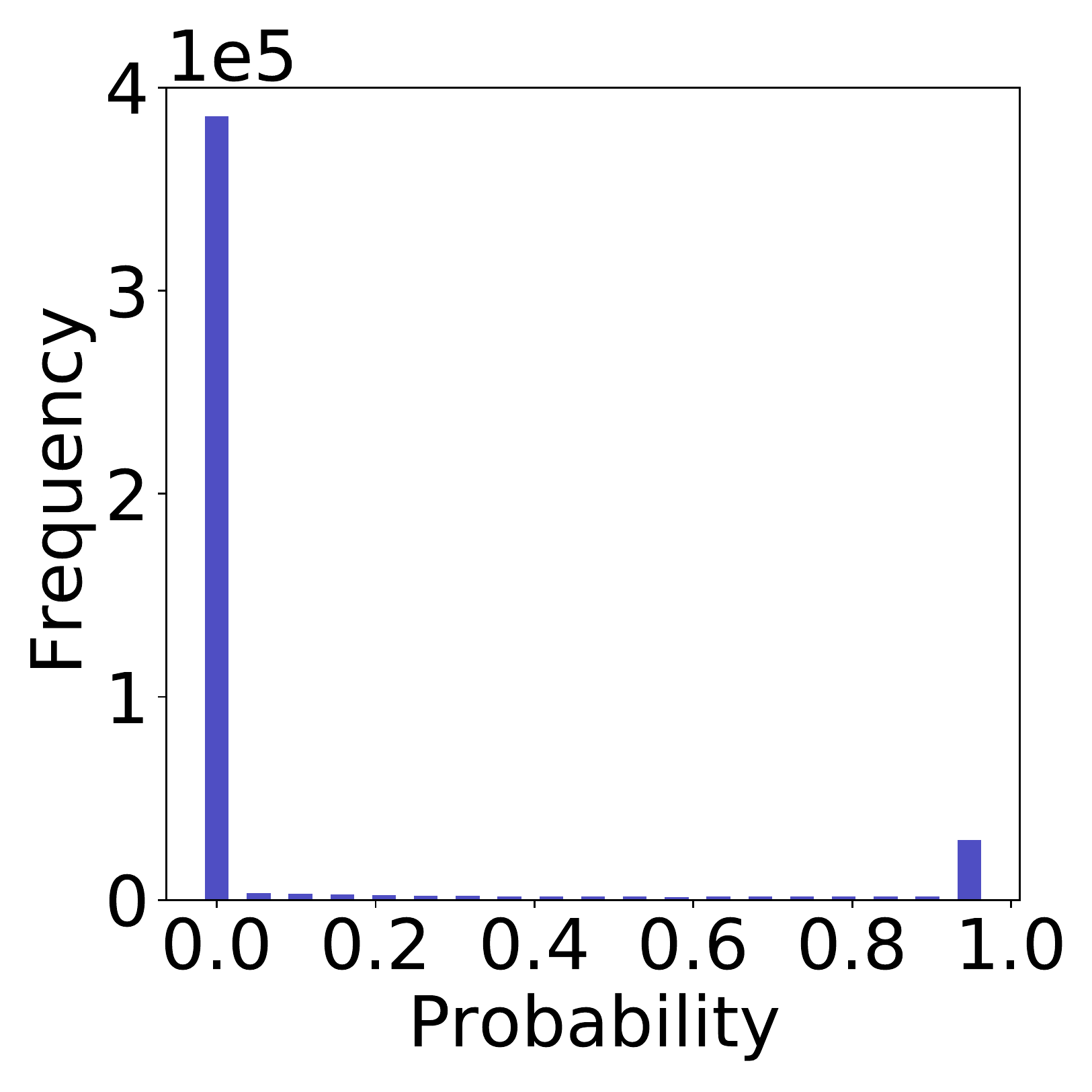}}
\hspace{0em}
\subfigure[epoch=30]{\label{epochfig:d}\includegraphics[width=24mm]{  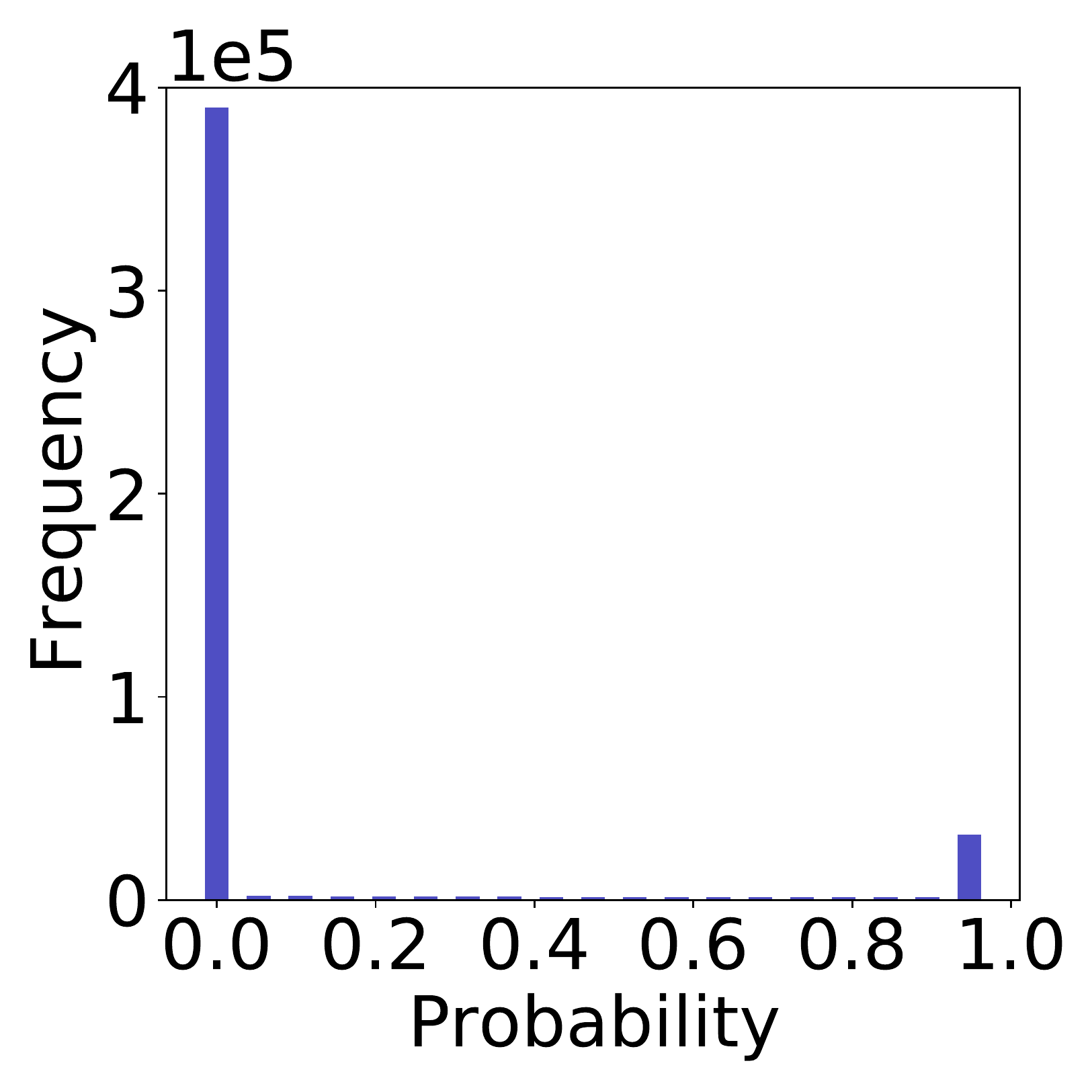}}
\hspace{0em}
\subfigure[epoch=40]{\label{epochfig:e}\includegraphics[width=24mm]{  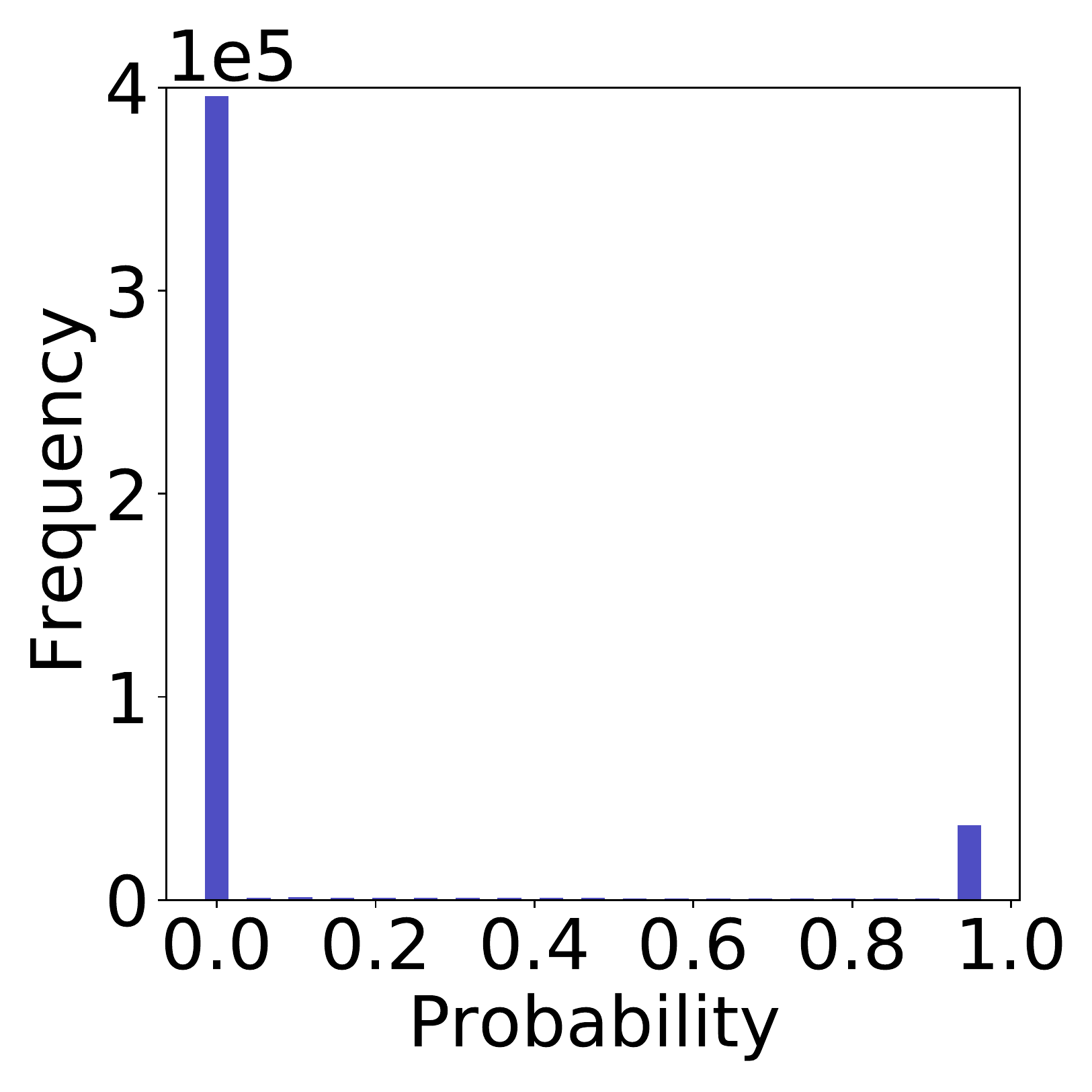}}
\hspace{0em}
\subfigure[epoch=50]{\label{epochfig:f}\includegraphics[width=24mm]{  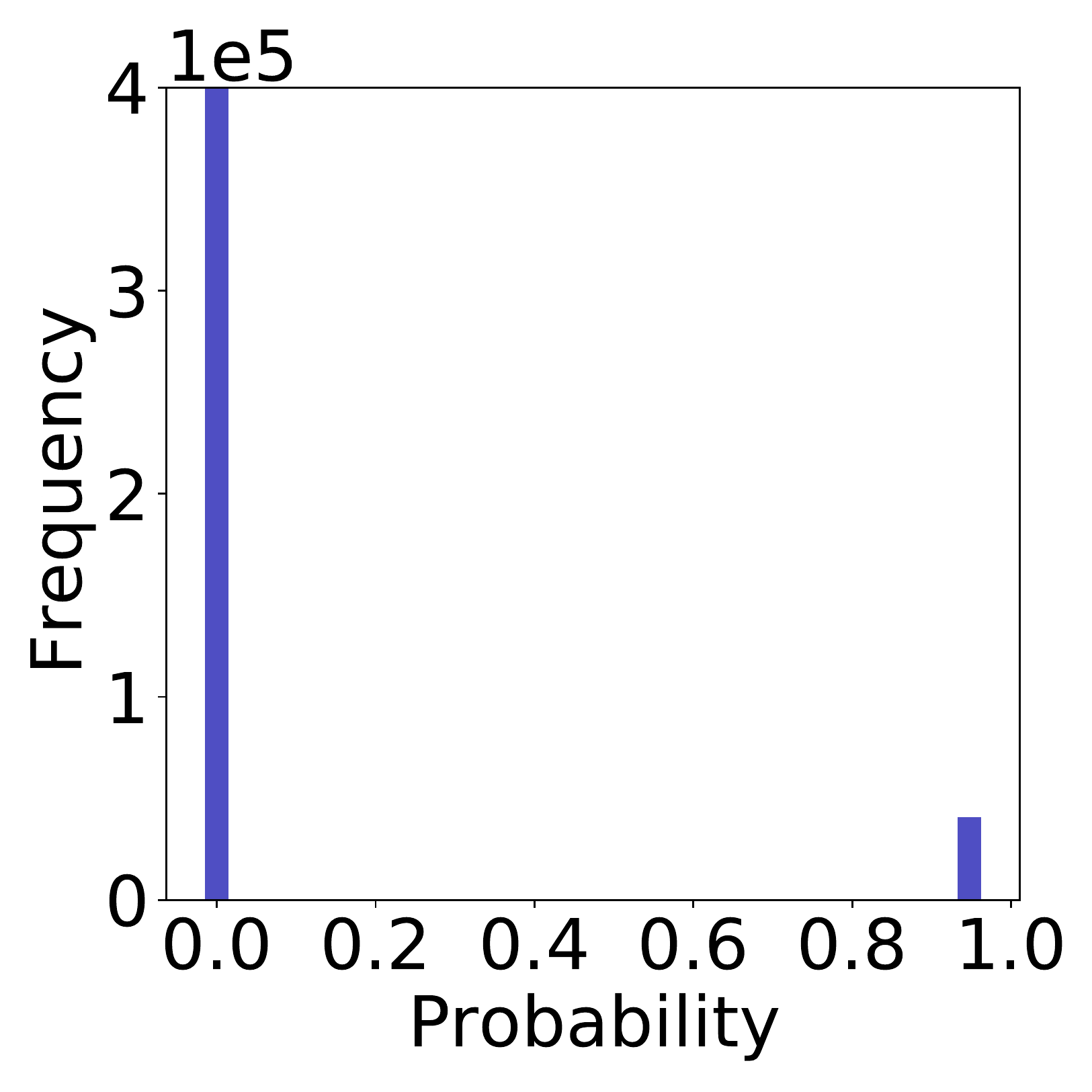}}
\hspace{0em}
\subfigure[epoch=100]{\label{epochfig:g}\includegraphics[width=24mm]{  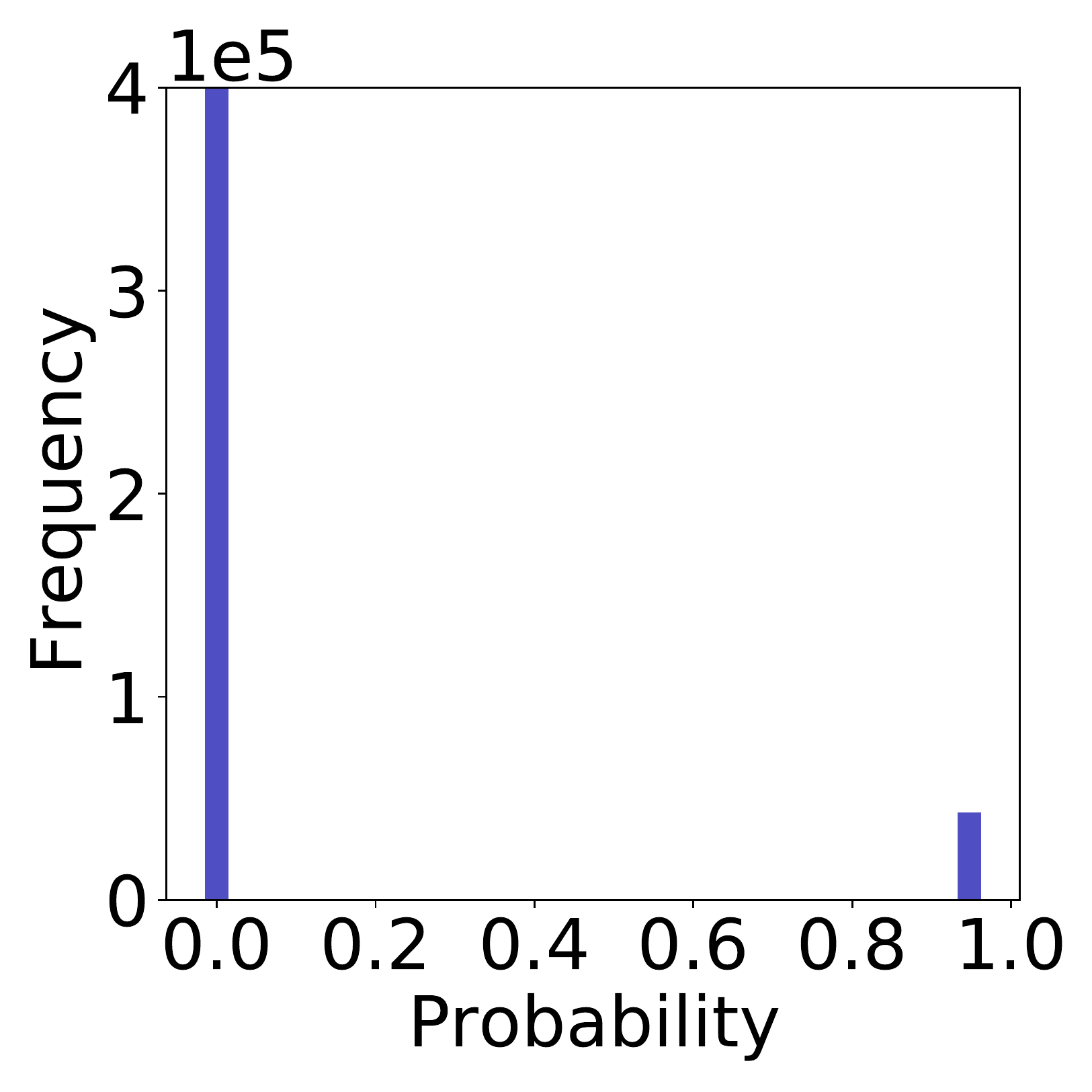}}
\vfill
\subfigure[epoch=1]{\label{epochfig:h}\includegraphics[width=24mm]{  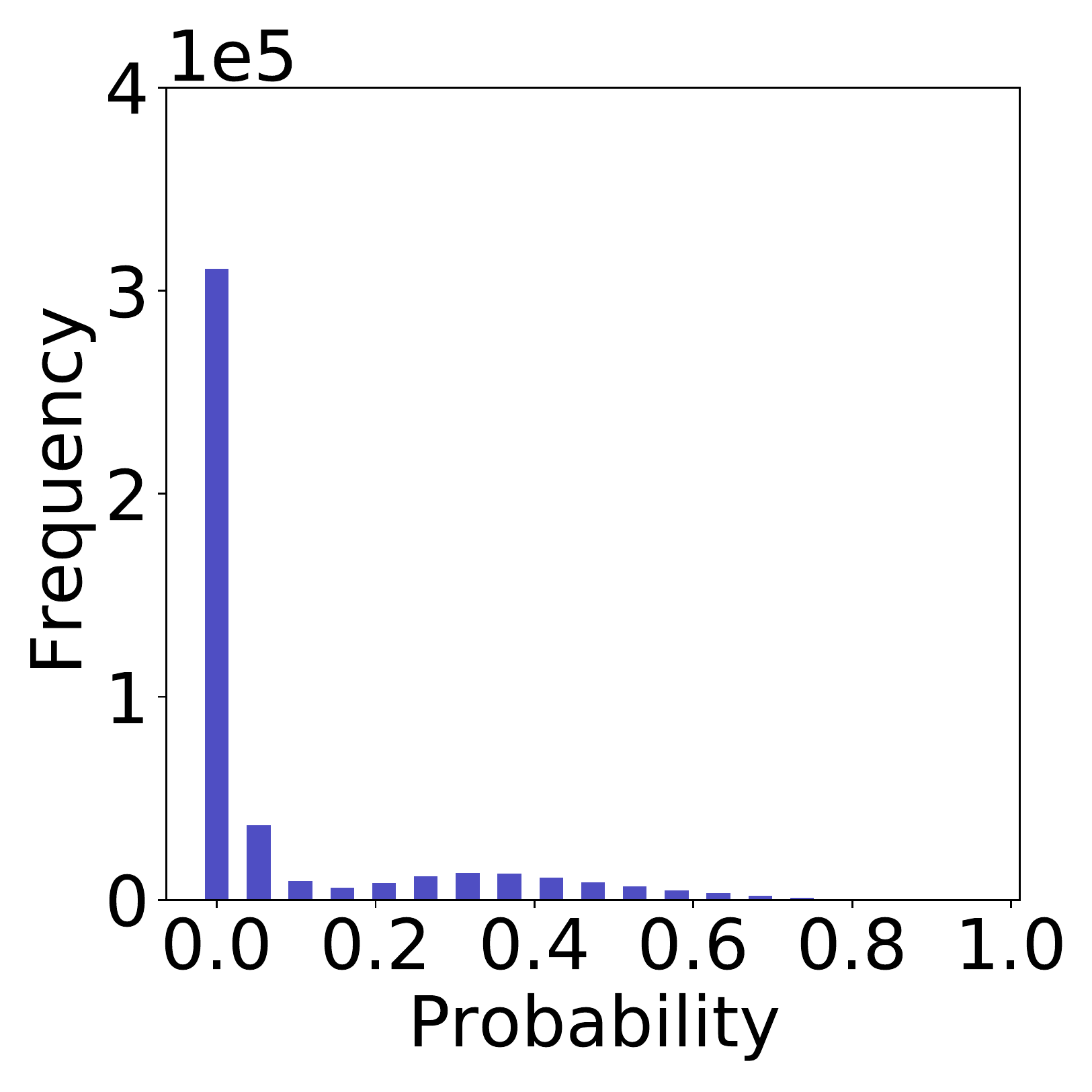}}
\hspace{0em}
\subfigure[epoch=10]{\label{epochfig:i}\includegraphics[width=24mm]{  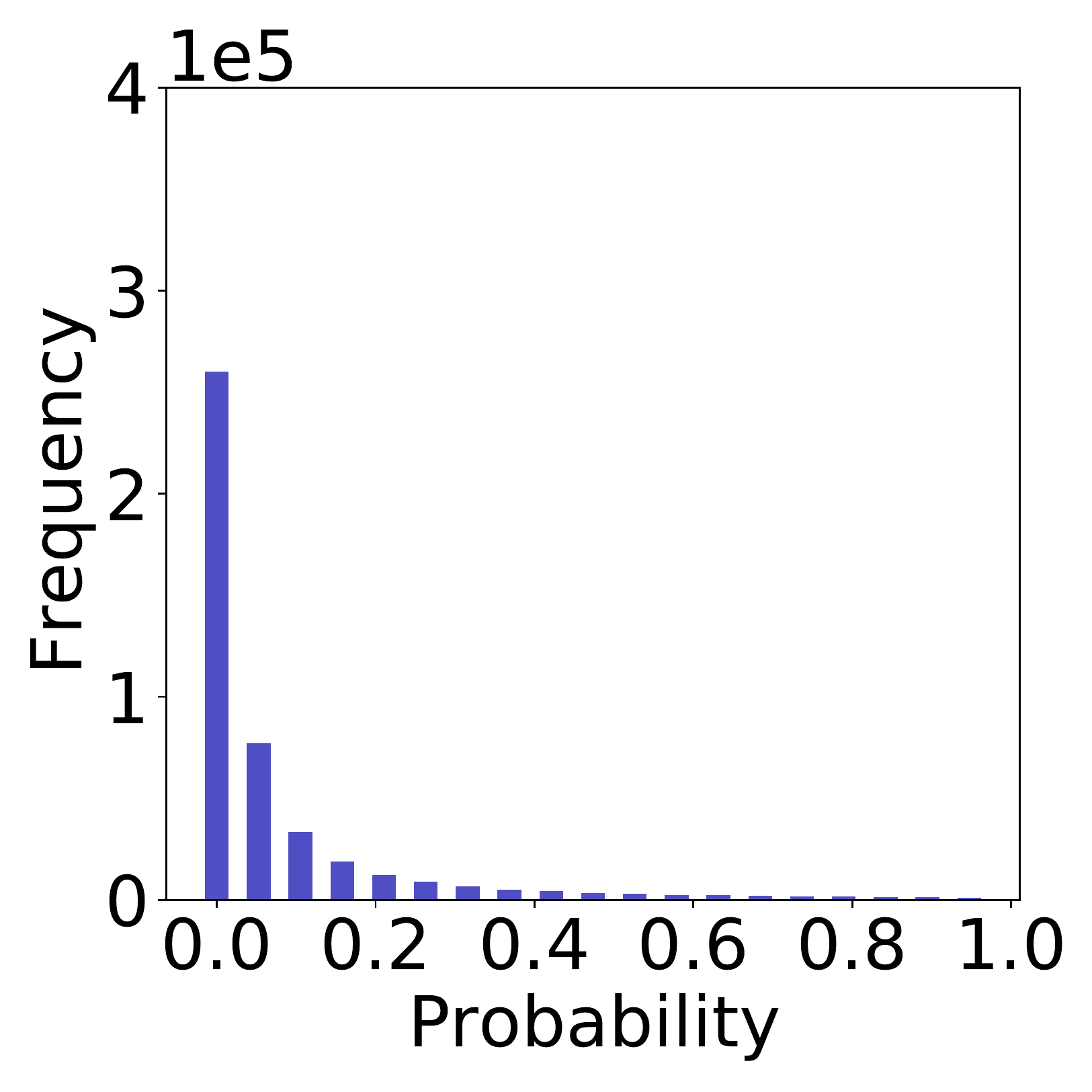}}
\hspace{0em}
\subfigure[epoch=20]{\label{epochfig:j}\includegraphics[width=24mm]{  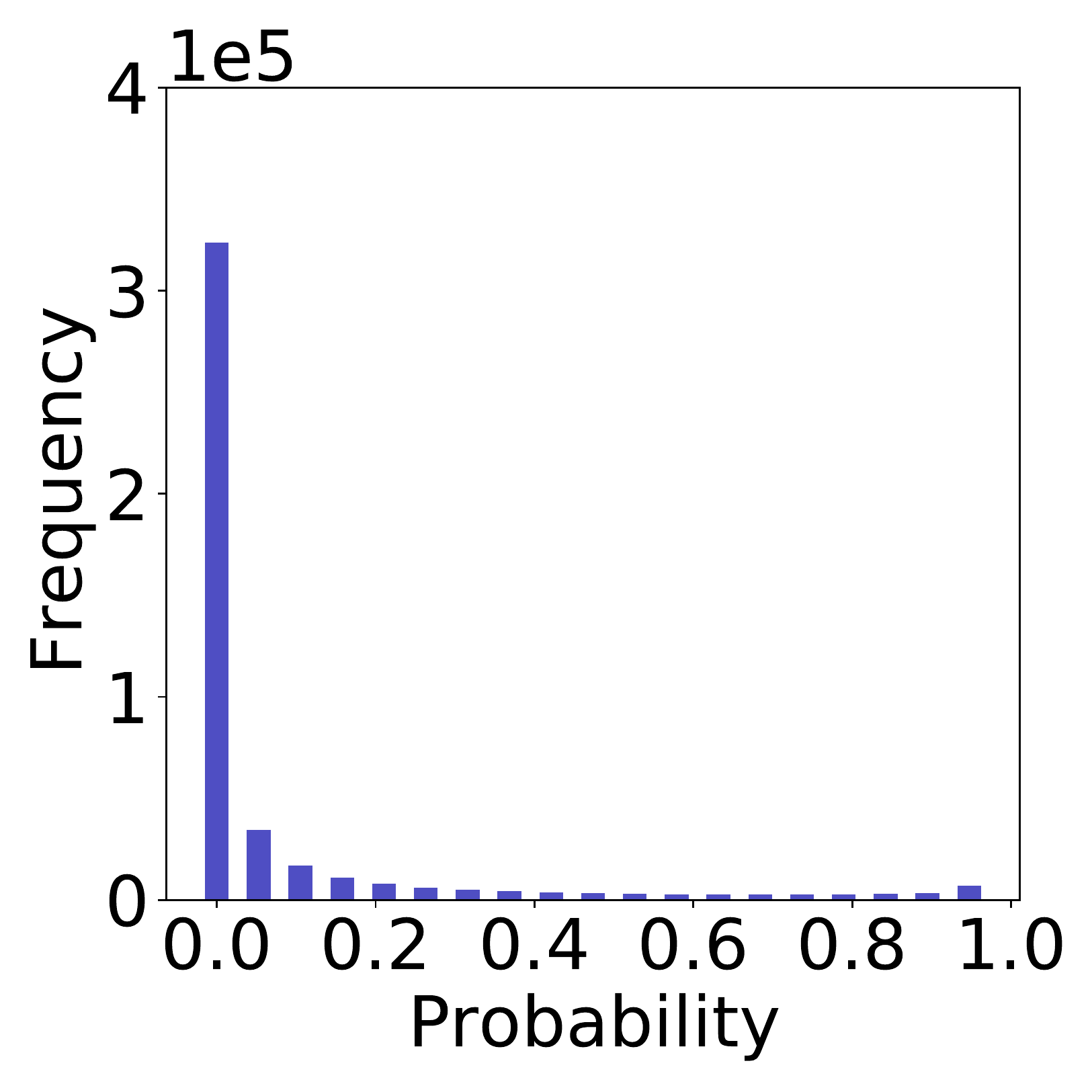}}
\hspace{0em}
\subfigure[epoch=30]{\label{epochfig:k}\includegraphics[width=24mm]{  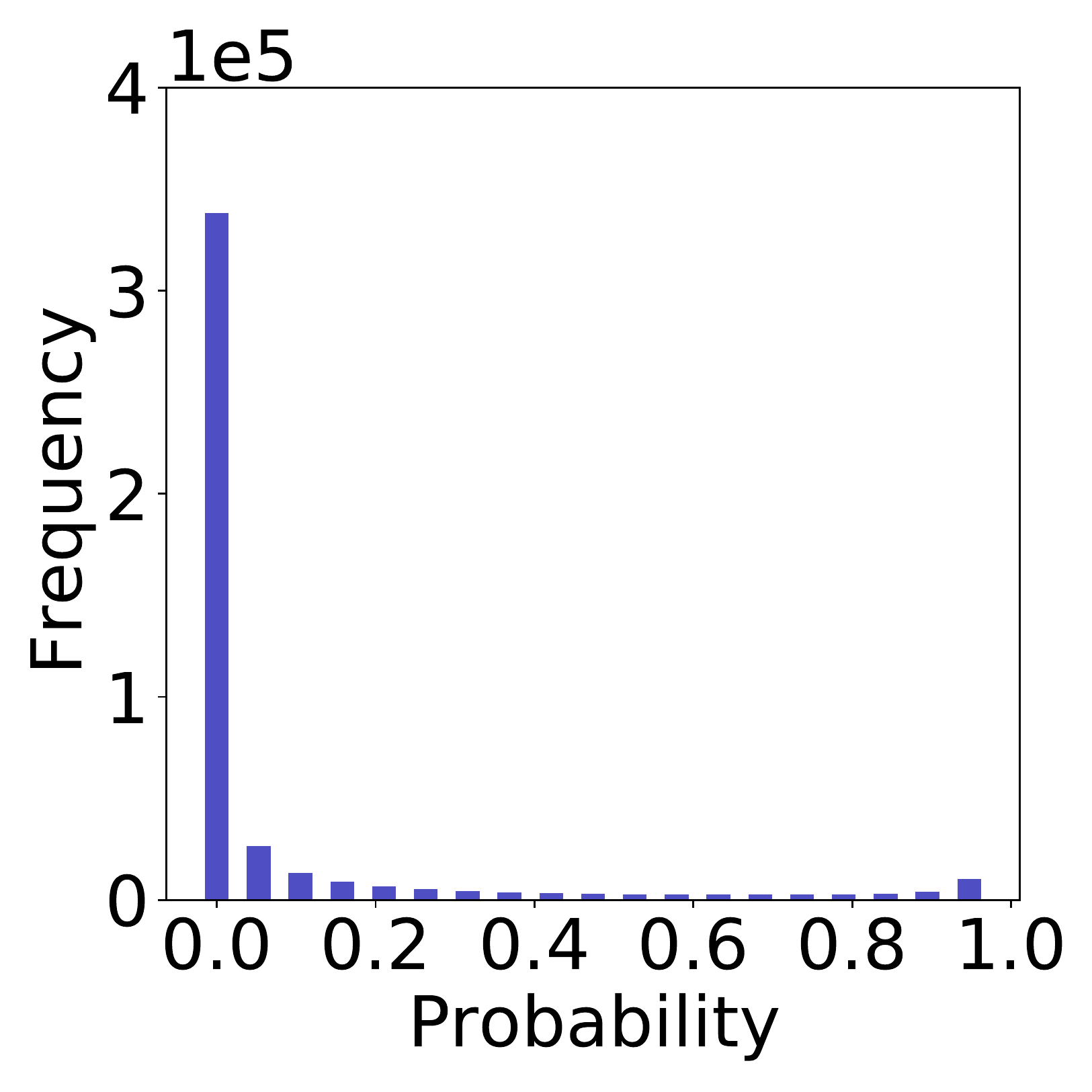}}
\hspace{0em}
\subfigure[epoch=40]{\label{epochfig:l}\includegraphics[width=24mm]{  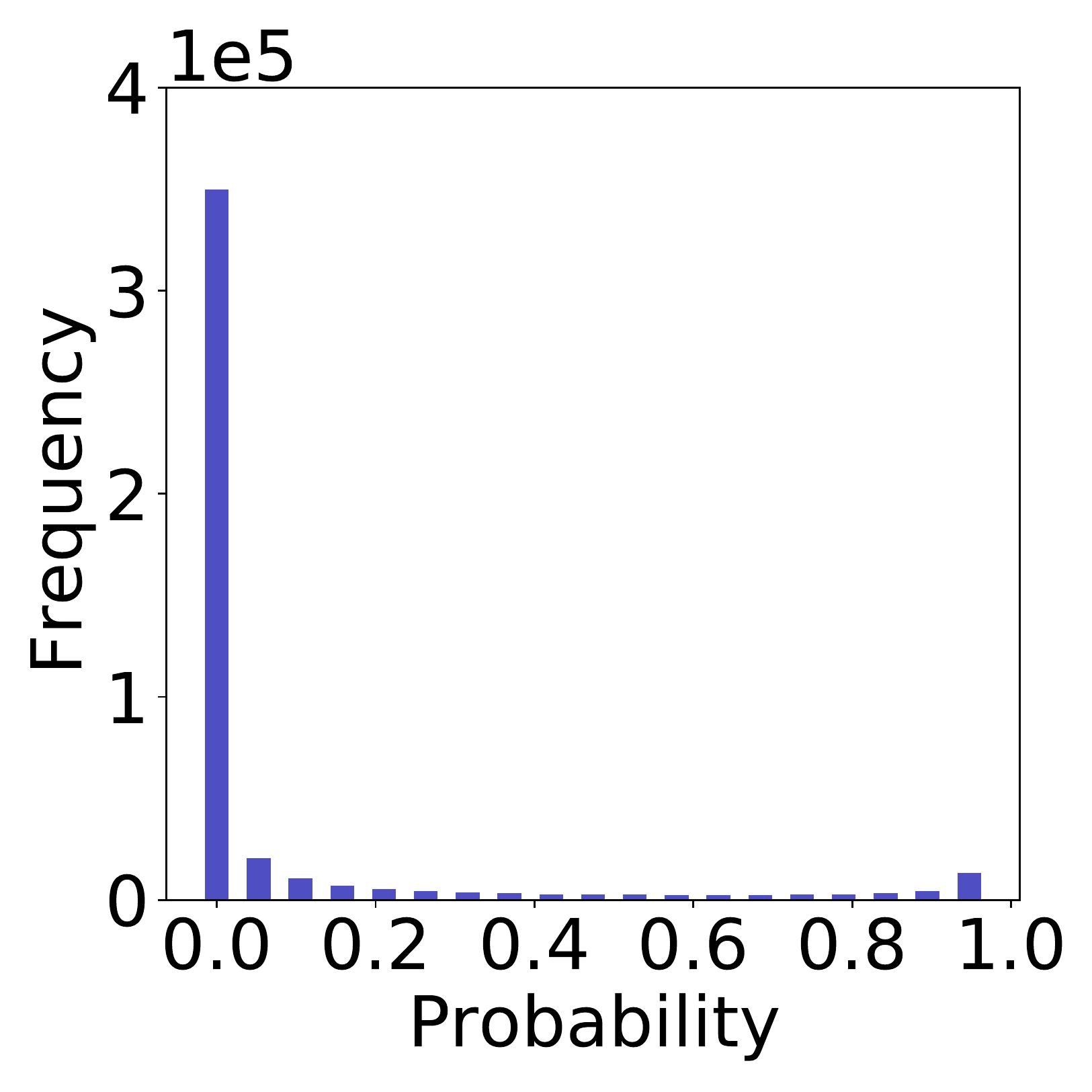}}
\hspace{0em}
\subfigure[epoch=50]{\label{epochfig:m}\includegraphics[width=24mm]{  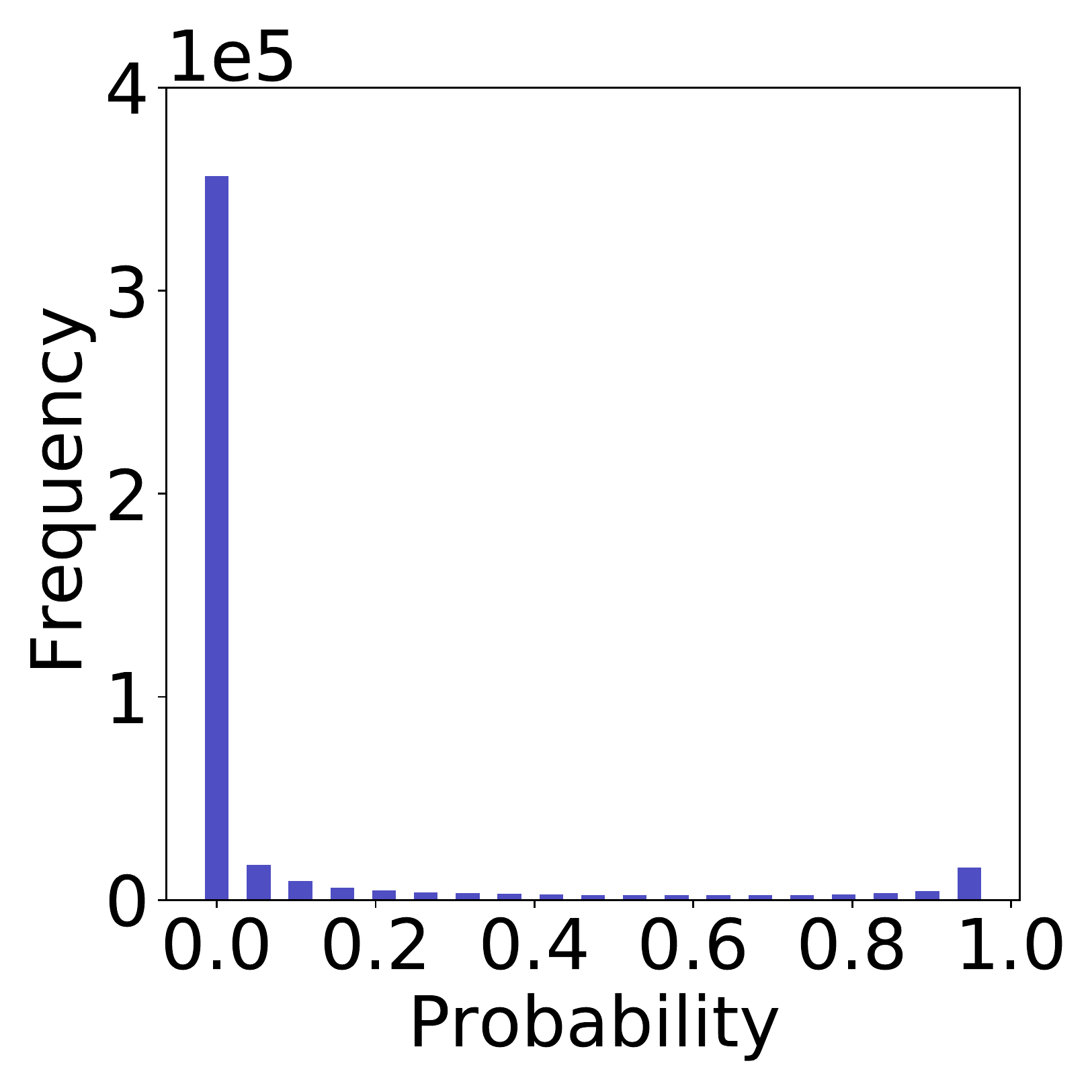}}
\hspace{0em}
\subfigure[epoch=100]{\label{epochfig:n}\includegraphics[width=24mm]{  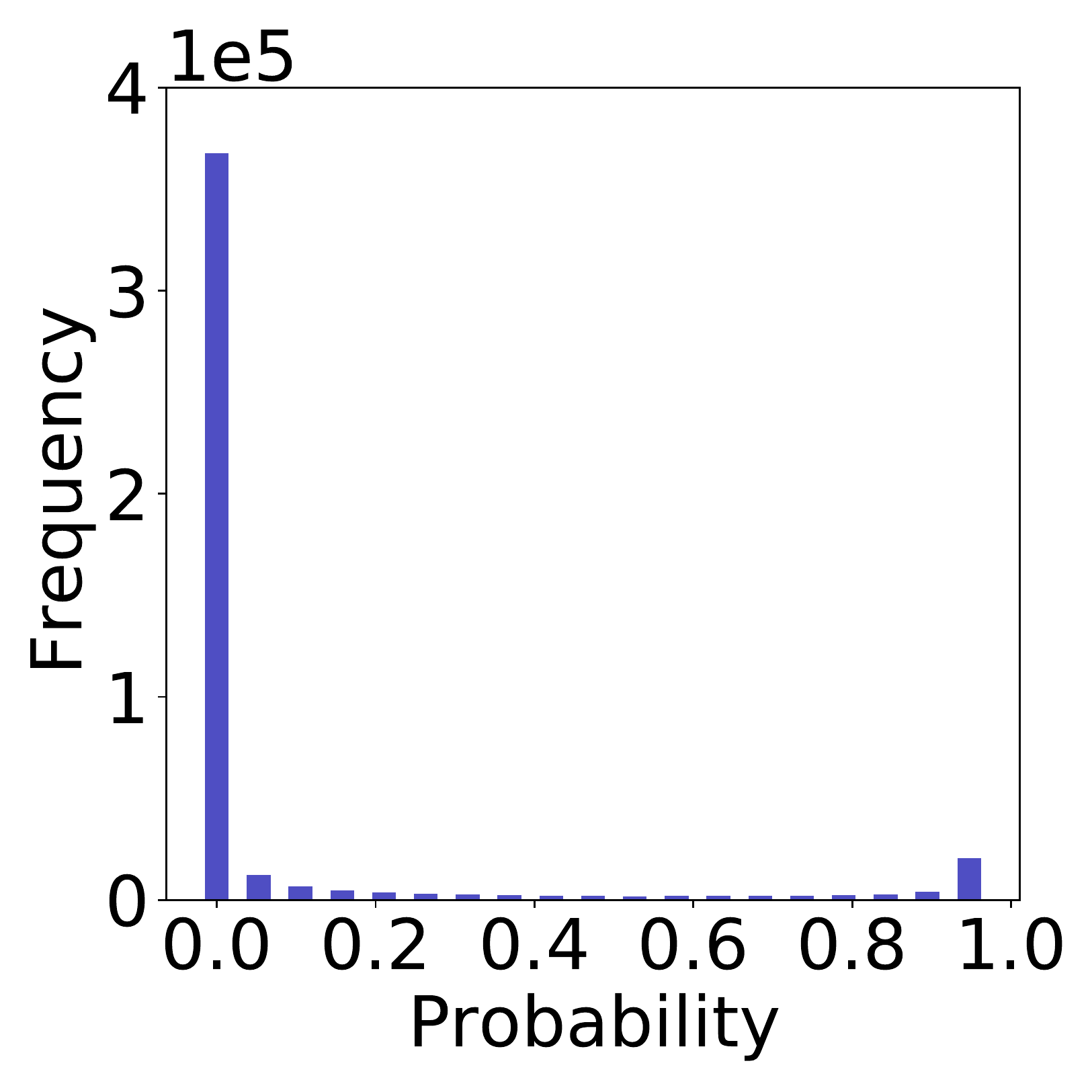}}
\caption{Numerical distribution of prediction values of VAT+\ADS on unlabeled training data. Top row shows the result on MNIST and the bottom row shows the result on CIFAR-10. The initialization of the network is used as their default. }
\label{fig:probabilities_frequency}
\end{figure*}

We further verify the mechanism of \ADS by recording the prediction frequency of unlabeled training data on MNIST and CIFAR-10 dataset. To this end, we evenly partition the prediction space into 10 intervals and let each of them serve as a bin to count the predictions whose values fall into it. Each bin is a half-closed interval except the last one. For example, the first bin is defined as $[0,0.1)$ and the last is $[09,1]$. The experiments are executed following the settings of Section~\ref{sec:5.3} where VAT serves as the base model.

Figure~\ref{fig:probabilities_frequency} presents the numerical distribution of prediction values over a series of epochs, where the maximum epoch is 100 because they are found not clearly altered after 100 epochs optimization. For MNIST, the prediction values are around 0.1 for every sample at the beginning. Then they gradually shift to other bins with the process of optimization. Eventually, almost all the predictions fall into the first and last bin, which means the predictions are distilled to be sufficiently certain. Regarding CIFAR-10, the results are a bit different. The initialization shows the model has some preference as mentioned in Section~\ref{sec:safety}. In addition, the predictions grouped into the last bin are relatively fewer than MNIST. This is because the discriminative information for CIFAR-10 images is harder to capture via unlabeled data. In other words, given a semi-supervised model and training data, \ADS tries to utilize the informed predictions and leaves the unreliable samples underfitted. 

\subsection{Ablation Study}
Our \ADS consistents of two main components:  \emph{sparsemax} for candidates selection, and \emph{Sharpening} for enhancing the sparse activations. In this section, we study their independent impact using CIFAR-10 with all but 4,000 labeled data. Particularly, we replace \ADS with different ablations, and combine each of them with VAT and FixMatch, respectively. 

Table~\ref{table:ablation} shows that: (1) \ADS with default parameter $r=2$ achieves the best performance, without extra efforts for fine-tuning regarding different datasets. (2) sparsemax solely imporves the basic SSL algorithms because it constrains the consistency loss focus on confusing classes only. (3) However, it hurts the performance when combining with other distilation approaches, such as Sparsemax+ME and Sparsemax+PL, becuase they fails to target on the most informative predictions.  Finally we conclude that sparsemax and Sharpening both contributes to \ADS.
\begin{table}[!ht]
\centering
\caption{Test error of ablation study on CIFAR-10 with 4,000 labels. Note that Sparsemax+ME does not apply to FixMatch.}
\begin{tabular}{ccc}
\toprule[1.3pt]
Ablation  & VAT   & FixMatch  \\ \midrule
None &$14.72$  &$5.92$\\
Sparsemax &$13.15$&$5.75$\\
Sparsemax+ME &$13.59$ &-\\
Sparsemax+PL &$14.26$&$5.75$\\
Sparsemax+NS &$13.02$&$5.44$\\
\ADS&$\bm{12.40}$ & $\bm{5.21}$\\
\bottomrule[1.3pt]
\end{tabular}
\label{table:ablation}
\end{table}


\subsection{Scarce Labeled Data Scenario}
One of the challenges in SSL is that in some scenarios only a limited number of labeled data are provided. FixMatch has shown its superiority to deal with the scarce labeled data thanks to its tailor-designed consistency loss. To confirm that the proposed \ADS also applies to this case, we conduct FixMatch-ADS with a small fraction of labeled data. Three datasets CIFAR-10, CIFAR-100, and SVHN with all but 40, 400, 40 labeled data are used following the setting of~\cite{sohn2020fixmatch}, and the results are shown in Table~\ref{table:small_labelled}. We observe that \ADS achieves better performance than the naive FixMatch across all the three experiment settings. In particular, the improvement on SVHN is significant (more than $5\%$), which demonstrates the efficacy of \ADS in the scenario with scare labeled data. 

\begin{table}[!ht] 
\centering
\caption{\label{table:small_labelled} Small fraction of labelled data  comparison on FixMatch. } 
\begin{tabular}{cccc}
\toprule[1.3pt]
Methods  & CIFAR-10 & CIFAR-100& SVHN  \\ \midrule[1pt]
FixMatch &${15.39}$  &${54.21}$ & $7.63$\\
FixMatch-\ADS&$\bm{14.84}$ &$\bm{52.83}$ & $\bm{2.37}$\\
\bottomrule[1.3pt]
\end{tabular}
\end{table}

\section{Discussion}\label{discussion}
In this paper, we have presented a new prediction distillation mechanism named ADaptive Sharpening (\ADS) which distills example-wise informed predictions to avert the risk of overconfident predictions. 
We discriminate \ADS from existing distillation strategies from the perspective of prediction uncertainty minimization. 
By integrating \ADS into SSL framework we discuss how \ADS affects other SSL components. The theoretical analyses present the valuable properties of \ADS which confirm our motivation.
Extensive experiments demonstrate the proposed \ADS forges a cornerstone for distillation-based SSL research. In future work, we would apply \ADS to more SSL tasks including different types of data~\cite{miyato2016adversarial,xu2019semisupervised} and unseen class scenarios~\cite{guo2020safe}.

\appendices
\section{Distillation comparison in the same probability space}\label{appendix:same_pro_space}
For the binary class case, softmax activation is known to degenerate into the logistic function. If $\bm{z}=(u,0)$, then $\text{softmax}_1(\bm{z})=\sigma(u):=(1+\exp(-u))^{-1}$. Thus, given  $\text{softmax}_1(\bm{z})=s$, we have $u=\ln \frac{s}{1-s}$. Meanwhile, according to the solution of sparsemax, for binary class case, it could be expressed as
\begin{equation}
    s':=\text{sparsemax}_1(u)=\left\{
    \begin{array}{ll}
     1, & { u>1 }\\
     (u+1)/2, & {-1 \le u \le 1}\\
     0, &  {u<-1}.
    \end{array}\right.
\end{equation}
Now we formulate sparsemax output $s'$ as a function of $s$
\begin{equation}\label{eq:s'_s}
    s'(s)=\left\{
    \begin{array}{ll}
     1, & { s>\frac{e}{e+1} }\\
     (\ln{\frac{s}{1-s}}+1)/2, & {\frac{1}{e+1} \le s \le \frac{e}{e+1}}\\
     0, &  {s<\frac{1}{e+1}},
    \end{array}\right.
\end{equation}
where $e$ is Euler's number. Note that Eq.~\eqref{eq:s'_s} is a binary case for Theorem~\ref{theorem1}.

Considering the distillation loss with the power $r=2$, we have the target probability
\begin{equation}\label{eq:t_s}
    t=\left\{
    \begin{array}{ll}
     1, & { s>\frac{e}{e+1} }\\
     t^\star, & {\frac{1}{e+1} \le s \le \frac{e}{e+1}}\\
     0, &  {s<\frac{1}{e+1}},
    \end{array}\right.
\end{equation}
where $t^\star$ is calculated as follows.
\begin{equation}
\begin{aligned}
    t^\star &=\frac{(\ln{\frac{s}{1-s}+1)^2/4}}{(\ln{\frac{s}{1-s}+1)^2/4}+(1-(\ln{\frac{s}{1-s}+1)/2})^2}\\
    &= \frac{1}{1+(\frac{\ln{\frac{e(1-s)}{s}}}{\ln{\frac{es}{1-s}}})^2} .
\end{aligned}
\end{equation}
It is easily verified that $t(s)$ is an odd function and $t^\star|_{s=0.5} = 0.5$. Based on Eq.~\eqref{eq:t_s}, we know the distillation loss (Eq.~\eqref{eq:J_d}) is nonzero iff ${\frac{1}{e+1} \le s \le \frac{e}{e+1}}$,
\begin{equation}\label{eq:full_obj}
\begin{aligned}
    \mathcal{J}_{\text{D}}(s) = & t\ln{t}+(1-t)\ln{(1-t)}\\
    &\underbrace{-t\ln{s'}-(1-t)\ln{(1-s')} }_{\text{reduced form}}.
\end{aligned}
\end{equation}
As $t$ will stop the gradients from propagation, let $s^\star: = (\ln{\frac{s}{1-s}}+1)/2$, and only the reduced form has the gradient
\begin{equation}
\begin{aligned}
    \nabla_s\mathcal{J}_{\text{D}} &= (-\frac{t}{s'} +\frac{1-t}{1-s'}) \nabla_s s' \\
    &= \frac{1}{2}(-\frac{t^\star}{s^\star} +\frac{1-t^\star}{1-s^\star})(\frac{1}{s}+\frac{1}{1-s}).
\end{aligned}
\end{equation}
The above derivatives are used to complete the plot of \ADS in Figure 3, and other distillation methods are displayed in a similar manner. 

\section{Proof for Corollary 1}\label{proof_corollary}
\noindent \textbf{Corollary 1.} \emph{For a $K$-way semi-supervised classification problem, the determinate predictions and negligible predictions for \ADS are masked out by the sample dependent threshold $\theta_1 \in [ \frac{e}{e+K-1},\frac{e}{e+1}]$ and $\theta_2 \in [ \frac{e^{\rho}}{\rho+e^{\rho}(K-\rho)},\frac{e^{\rho}}{\rho+e^{\rho}}]$ in the corresponding softmax output space, respectively, where $e$ is Euler number and $\rho$ is the population of non-zero predictions.}

\begin{proof}
A prediction $p$ is said determinate for \ADS if $p=p_{(1)}$ and $p_{(1)}\ge ep_{(2)}$. Obviously, for a $K$-way classification the maximum $\theta_1$ is picked if $p_{(3)} = p_{(4)} = ... =p_{(K)}= 0$. Combining with the equality $\sum_kp_{(k)}=1$, we have $\argmax {\theta_{1}}=\frac{e}{e+1}$. Meanwhile, the minimum $\theta_1$ is picked if $p_{(2)} = p_{(3)} = ... = p_{(K)}$. Thus we have $\argmin {\theta_{1}}=\frac{e}{e+K-1}$. 

Let $\rho$ denote the population of non-zero outputs given a logits vector $\bm{z}$. According to the solution of sparsemax, if $\rho<K$, we have the inequality
\begin{equation}\label{eq:co2_1}
    1+ (\rho+1) z_{(\rho+1)} \le \sum_{j=1}^{\rho+1} z_{(j) }.
\end{equation}
Following the derivation of Eq.~\eqref{eq:theorem3}, we rewrite Eq.~\eqref{eq:co2_1} as
\begin{equation}
\begin{aligned}
    &1+(\rho+1) \ln Cp_{(\rho+1)} \le \sum_{j=1}^{\rho+1} \ln Cp_{(j)}\\ \Rightarrow & p_{(\rho+1)} \le (\frac{1}{e}\prod_{j=1}^{\rho}p_{(j)})^{\frac{1}{\rho}}.
\end{aligned}
\end{equation}
Similar to the proof to $\theta_1$, the maximum of $\theta_2$ is derived if $p_{(\rho+2)}=...=p_{(K)}=0$, and we have 
\begin{equation}
p_{(\rho+1)} =  \frac{e^{\rho}(1-p_{(\rho+1)})}{\rho},
\end{equation}
which suggests $ \argmax {\theta_2} = \frac{e^{\rho}}{\rho+e^{\rho}}$. The minimum of $\theta_2$ is obtained if $p_{(\rho+1)}=p_{(\rho+2)}=...=p_{(K)}$. That is
\begin{equation}
p_{(\rho+1)} = \frac{e^{\rho}(1-(K-\rho)p_{(\rho+1)})}{\rho},
\end{equation} 
which suggests $ \argmax {\theta_2} = \frac{e^{\rho}}{\rho+e^{\rho}(K-\rho)}$.

The above bounds complete the proof to the range of threshold  for $\theta_1$ and $\theta_2$ in \ADS in terms of softmax output space. 
\end{proof}



\ifCLASSOPTIONcaptionsoff
  \newpage
\fi



%



\bibliographystyle{IEEEtran}
\bibliography{IEEEabrv}
%





\end{document}